%% file: main.tex
\documentclass[10pt,twocolumn,journal]{IEEEtran}
\usepackage{setspace}
\usepackage{amsmath}
\usepackage{amsthm}
\usepackage{algorithmic}
\usepackage{multirow}
\usepackage{bigstrut}
\usepackage{graphicx}
\usepackage{hhline}
\usepackage{cite}
\usepackage{algorithm}
\usepackage{amssymb}
\usepackage{epsfig}
\usepackage{array}
\usepackage{subfigure}
\usepackage{bm}
\usepackage{enumerate}
\usepackage{arydshln}
\usepackage{extarrows}
\usepackage{color}
\usepackage{stackengine}
\usepackage[colorlinks,linkcolor=black,anchorcolor=black,citecolor=black,urlcolor=black]{hyperref}

\newtheorem{theorem}{Theorem}

\theoremstyle{definition}

\theoremstyle{remark}

\newcommand\given[1][]{#1\vert}


\begin{document}
%
\title{Graph Laplacian Regularization for Image Denoising: Analysis in the Continuous Domain}

\author{Jiahao Pang,~\IEEEmembership{Member,~IEEE},
Gene Cheung,~\IEEEmembership{Senior Member,~IEEE}

\begin{small}

\thanks{J. Pang conducted the research while at the Department of Electronic and Computer Engineering, The Hong Kong University of Science and Technology, Hong Kong. He is currently with SenseTime Group Limited, Hong Kong (e-mail: pangjiahao@sensetime.com).}

\thanks{G. Cheung is with the National Institute of Informatics, Graduate University for Advanced Studies, Tokyo 113-0033, Japan (e-mail: cheung@nii.ac.jp).}

\thanks{This work was partially supported by JSPS Grant-in-Aid for Challenging Exploratory Research (15K12072).}

\end{small}
}

\maketitle

\begin{abstract}
Inverse imaging problems are inherently under-determined, and hence it is important to employ appropriate image priors for regularization. 
One recent popular prior---the graph Laplacian regularizer---assumes that the target pixel patch is smooth with respect to an appropriately chosen graph. 
However, the mechanisms and implications of imposing the graph Laplacian regularizer on the original inverse problem are not well understood. 
To address this problem, in this paper we interpret neighborhood graphs of pixel patches as discrete counterparts of Riemannian manifolds and perform analysis in the continuous domain, providing insights into several fundamental aspects of graph Laplacian regularization for image denoising. 
Specifically, we first show the convergence of the graph Laplacian regularizer to a continuous-domain functional, integrating a norm measured in a locally adaptive metric space. 
Focusing on image denoising, we derive an optimal metric space assuming non-local self-similarity of pixel patches, leading to an optimal graph Laplacian regularizer for denoising in the discrete domain. 
We then interpret graph Laplacian regularization as an anisotropic diffusion scheme to explain its behavior during iterations, e.g., its tendency to promote piecewise smooth signals under certain settings. 
To verify our analysis, an iterative image denoising algorithm is developed. 
Experimental results show that our algorithm performs competitively with state-of-the-art denoising methods such as BM3D for natural images, and outperforms them significantly for piecewise smooth images.
\end{abstract}


%
\IEEEpeerreviewmaketitle

\section{Introduction}
\label{sec:intro}
\input{intro}

\section{Related Work}
\label{sec:related}
\input{related}

\section{Interpreting Graph Laplacian Regularizer in the Continuous Domain}
\label{sec:interp_xlx}
\input{interp_xlx}

\section{Optimal Graph Laplacian Regularizer for Image Denoising}
\label{sec:optim_xlx}
\input{optim_xlx}

\section{Analyzing Graph Laplacian Regularization by Anisotropic Diffusion}
\label{sec:diffu_xlx}
\input{diffu_xlx}

\section{Algorithm Development}
\label{sec:alg}
\input{alg}

\section{Experimentation}
\label{sec:results}
\input{results}
\section{Conclusion}
\label{sec:conclude}
\input{conclude}

\appendices

\section{\small{Proof of Theorem 1}}
\label{app:conv_proof}
\input{conv_proof}

\section*{Acknowledgment}
\addcontentsline{toc}{section}{Acknowledgment}
The authors thank Prof. Antonio Ortega of University of Southern California for his insightful comments, which improved the technical quality of our work.




\bibliographystyle{IEEEtran}
\bibliography{./IEEEabrv_mod,./ref}

\input{bio}

\end{document}

%% file: intro.tex
\subsection{Motivation}
\IEEEPARstart{I}{n} an inverse imaging problem, one seeks the original image given one or more observations degraded by corruption, such as noise, blurring or lost components (in the spatial or frequency domain). 
An inverse problem is inherently under-determined, and hence it is necessary to employ image priors to regularize it into a well-posed problem. 
Proposed image priors in the literature include total variation (TV) \cite{rudin92}, sparsity prior \cite{elad06} and autoregressive prior \cite{zhang08}. 
Leveraging on the recent advances in \textit{graph signal processing} (GSP) \cite{shuman13,sandryhaila13}, a relatively new prior is the \emph{graph Laplacian regularizer}, which has been shown empirically to perform well, despite its simplicity, in a wide range of inverse problems, such as denoising \cite{kheradmand13,hu13,liu14}, super-resolution \cite{hu14,wang14}, deblurring~\cite{kheradmand14}, de-quantization of JPEG images \cite{liu15,hu16,liu17} and bit-depth enhancement \cite{wan14,wan16}.
We study the mechanisms and implications of graph Laplacian regularization for inverse imaging problems in this paper.

Different from classical digital signal processing with regular data kernels, GSP assumes that the underlying data kernel is structured and described by a graph. 
Though a digital image lives on a 2D grid, one can nonetheless view pixels as vertices $\cal V$ connected via edges $\cal E$ with weights $\bf A$ on a neighborhood graph $\cal G$.  
Edge weights $\bf A$ model the similarity/affinity between pairs of pixels. 
Such a graph construction enables us to interpret an image (or image patch) $\bf u$ as a graph-signal residing on a finite graph $\mathcal{G}(\cal V, \cal E, \bf A)$. 

A graph Laplacian regularizer assumes that the original image (patch) ${\bf u}$ is \emph{smooth} with respect to a defined graph $\cal{G}$.
Specifically, it states that the ground-truth image (patch) ${\bf u}$ in vector form should induce a small value $S_{\cal G}({\bf u})={\bf u}^{\rm T}{\bf L}{\bf u}$, where ${\bf L}$ is the \emph{graph Laplacian matrix} of graph $\cal{G}$. 
Thus, for instance, to denoise an observed pixel patch ${\bf{z}}_0$, one can formulate the following unconstrained quadratic programming (QP) problem:
\begin{equation}\label{eq:prob_discrete}
{{\bf{u}}^ \star } = \mathop {\arg \min }\limits_{\bf{u}} \left\| {{\bf{u}} - {{\bf{z}}_0}} \right\|_2^2+ \tau\bm\cdot{{\bf{u}}^{\rm{T}}}{\bf{Lu}},
\end{equation}
where $\tau$ is a weighting parameter.
This is a straightforward formulation combining the prior term $S_{\cal G}({\bf u})$ with an $\ell_2$-norm fidelity term computing the difference between the noisy observation ${\bf{z}}_0$ and the denoised patch $\mathbf{u}$.
For a \emph{fixed} ${\bf L}$, \eqref{eq:prob_discrete} admits a closed-form solution linear to ${\bf z}_0$, {\it i.e.,} ${{\bf{u}}^ \star }={\left( {{\bf{I}} + \tau {\bf{L}}} \right)^{ - 1}}\bm\cdot{{\bf{z}}_0}$. We will develop a graph Laplacian matrix ${\bf L}$ that depends on ${\bf z}_0$, leading to a \emph{non-linear} filtering.

Despite the simplicity and success of graph Laplacian regularization in various inverse imaging problems---with significant gain over state-of-the-art methods for piecewise smooth images like depth maps \cite{wang14,hu13,hu14} and \cite{hu16}---there is still a lack of fundamental understanding of how it works and why it works so well. 
In particular:
\begin{enumerate}[(i)]
\item{How does the graph Laplacian regularizer promote a correct solution to restore a corrupted image effectively?}
\item{What is the optimal graph, and hence the optimal graph Laplacian regularizer, for inverse imaging?}
\item{Why does the graph Laplacian regularization perform particularly well on piecewise smooth images?}
\end{enumerate}
%
%
\subsection{Our Contributions}
In this paper, by viewing neighborhood graphs of pixel patches as discrete counterparts of Riemannian manifolds \cite{hein06,ting10} and analyzing them in the continuous domain, we provide answers to the aforementioned open questions:
\begin{enumerate}[(i)]

\item We first show the convergence of the graph Laplacian regularizer to the anisotropic Dirichlet energy \cite{flucher99}---a continuous-domain functional integrating a norm measured in a locally adaptive \emph{metric space}. Analysis of this functional reveals what signals are being discriminated and to what extent, thus explaining the mechanism of graph Laplacian regularization for inverse imaging.

\item Focusing on the most basic inverse imaging problem---image denoising---we derive an \emph{optimal} metric space by assuming non-local self-similarity of image patches, leading to the computation of optimal edge weights and hence the optimal graph Laplacian regularizer for denoising in the discrete domain.

\item We interpret graph Laplacian regularization as an \emph{anisotropic diffusion} scheme in the continuous domain to understand its behavior during iterations. Our analysis shows that graph Laplacian regularization not only smooths but may also sharpens the image, which explains its tendency to promote piecewise smooth images under specific settings. We also delineate the relationship between graph Laplacian regularization and several existing works such as TV \cite{rudin92} for denoising.

\end{enumerate}

To demonstrate the usefulness of our analysis, we develop an iterative algorithm called \emph{optimal graph Laplacian regularization} (OGLR) for denoising. 
Experimental results show that our OGLR algorithm performs competitively with state-of-the-art denoising methods such as BM3D \cite{dabov07} for natural images, and outperforms them significantly for piecewise smooth images.

Our paper is organized as follows. 
We review related works in Section \ref{sec:related}. 
In Section \ref{sec:interp_xlx}, we analyze the graph Laplacian regularizer in the continuous domain. 
With the insights obtained in Section\;\ref{sec:interp_xlx}, we derive in Section \ref{sec:optim_xlx} the optimal graph Laplacian regularizer for image denoising. 
In Section\;\ref{sec:diffu_xlx}, graph Laplacian regularization is interpreted as an anisotropic diffusion scheme in the continuous domain to explain its behavior. 
Then an iterative denoising algorithm is developed in Section\;\ref{sec:alg}. 
The experimental results and conclusions are presented in Section\;\ref{sec:results} and \ref{sec:conclude}, respectively.

%% file: related.tex
We first review recent works that employ the graph Laplacian regularizer or its variants
as priors for inverse imaging. 
We then review several representative works in image denoising. 
Finally, we review some works that relate graphs to Riemannian manifolds and anisotropic diffusion, and works that recover images with Riemannian metrics.

\subsection{Graph-Based Smoothness Prior for Inverse Imaging}\label{ssec:gsp_review}

\subsubsection{Graph Laplacian regularization} 

In \cite{liu14}, Liu~{\it et~al.} applied multi-scale graph Laplacian regularization for impulse noise removal. 
Using a variant of the normalized graph Laplacian, Kheradmand~{\it et~al.}~\cite{kheradmand14,kheradmand13} developed a framework for image deblurring and denoising. 
In \cite{hu14}, Hu~{\it et~al.} employed graph Laplacian regularization for joint denoising and super-resolution of generalized piecewise smooth images. 
While these works show good performance in different inverse problems using graph Laplacian regularization, they lack a clear exposition of why the graph Laplacian approach works---a missing link we provide in our study.
Note that this paper is a non-trivial extension of our previous works \cite{pang14,pang15}; we provide here a more thorough analysis of graph Laplacian regularization and interpret it as anisotropic diffusion for further insights and connections to previous works like TV~\cite{rudin92}.

\subsubsection{Other smoothness priors} 

In \cite{wang14}, Wang~{\it et~al.} employed a high-pass graph filter for regularization, and performed super-resolution on depth images. 
By assuming sparsity in the graph frequency domain, Hu~{\it et~al.} \cite{hu13} developed the non-local graph-based transform (NLGBT) algorithm for depth image denoising and achieved good performance. 
Other graph-based smoothness priors include the discrete $p$-Dirichlet energy \cite{elmoataz08}, graph total variation \cite{tovsic10}, etc. 
In contrast, our work focuses on the analysis and application of the graph Laplacian regularizer for image denoising.

It is shown in \cite{hu13,hu14} and \cite{wang14} that graph-based smoothness priors perform particularly well when restoring piecewise smooth images, {\it e.g.}, depth maps. 
Nevertheless, none of these works provide a theoretically justified explanation for this remarkable result. We provide this missing link in Section~\ref{sec:diffu_xlx}.

\subsection{Image Denoising}

Image denoising is a basic yet challenging problem that has been studied for decades.
As mentioned by Buades~{\it et~al.}~\cite{buades11}, denoising is essentially achieved by {\it averaging}. Depending on whether such averaging is carried out {\it locally} or {\it non-locally}, denoising algorithms can be classified into two categories.

\subsubsection{Local methods} 

Rudin~{\it et~al.}~\cite{rudin92} proposed to minimize the TV norm subject to constraints of the noise statistics. 
In \cite{perona90}, Perona~{\it et~al.} proposed anisotropic diffusion (Perona-–Malik diffusion) to remove noise while preserving edges. 
Inspired by \cite{strong96}, we will show that both TV denoising and Perona\--Malik diffusion have deep connections to graph Laplacian regularization. 
The recent work \cite{lefkimmiatis15} by \mbox{Lefkimmiatis}~{\it et~al.} employs a continuous functional called structure tensor total variation (STV) to penalize the eigenvalues of the structure tensor of a local neighborhood. Though our metric space is also closely related to the notion of structure tensor, we compute it optimally based on a set of non-local similar patches and the noise variance. 
Other local methods include bilateral filtering (BF) \cite{tomasi98}, wavelet thresholding \cite{donoho95}, the locally adaptive regression kernel (LARK) \cite{takeda07}, etc. 
In general, local methods are simpler but are inferior to non-local methods.

\subsubsection{Non-local methods} 

Buades~{\it et~al.} \cite{buades05} proposed \textit{non-local means} (NLM) denoising, assuming that similar image patches recur non-locally throughout an image. Such a {\it self-similarity} assumption has proven effective and has been adopted in many subsequent proposals.
One state-of-the-art method, \textit{block-matching 3-D} (BM3D)~\cite{dabov07}, performs shrinkage in the 3-D transform domain and Wiener filtering on the grouped similar patches. 
Elad~{\it et~al.} \cite{elad06} proposed K-SVD denoising, which seeks sparse codes to describe noisy patches using a dictionary trained from the whole noisy image. 
Based on a performance bound of image denoising \cite{chatterjee10}, Chatterjee~{\it et~al.}\cite{chatterjee12} proposed patch-based locally optimal Wiener filtering (PLOW).
To seek sufficiently similar patches, Talebi~{\it et~al.}~\cite{talebi14} developed a paradigm which enables existing denoising filters to collect similar patches from the whole image. 
While our work is also a non-local method, we construct an \emph{optimal} graph Laplacian regularizer from the non-local similar patches. 
Further, we also analyze its behavior using the notion of anisotropic diffusion, {\it e.g.}, its tendency to promote piecewise smooth signals under different settings. 
\emph{While we have developed an algorithm for proof of concept that is competitive with state-of-the-art image denoising schemes, we stress that the main objective of our work is to provide fundamental insights into the graph Laplacian regularizer, which we believe are useful for other inverse imaging problems as well.}

%
%


\subsection{Other Related Work}\label{ssec:related_other}
\subsubsection{Graph and Riemannian manifold} 
There exists some works linking operations on graphs to their manifold counterparts. 
In \cite{ting10,belkin05} and \cite{hein07}, the authors showed convergence of the graph Laplacian operator to the continuous Laplace-–Beltrami operator. 
In \cite{hein06}, Hein further showed convergence of the graph Laplacian regularizer to a functional for H\"{o}lder functions on Riemannian manifolds. 
Based on \cite{hein06}, our work focuses on graphs accommodating 2D image signals and proves the convergence of the graph Laplacian regularizer to a functional for continuous image signals. 
Our convergence result is non-trivial since it requires a conversion of the Laplace--Beltrami operator for general Riemannian manifolds to a simpler functional for 2D images.

\subsubsection{Graph and anisotropic diffusion}
Anisotropic diffusion smooths images in an edge-aware manner\cite{perona90,weickert98}. 
To execute a continuous-domain anisotropic diffusion forward in time, one may first discretize it with neighborhood graphs as done in \cite{gilboa07}. 
Some works have also proposed to diffuse discrete signals on graphs directly, {\it e.g.}, \cite{zhang08diff} and \cite{segarra15}. 
Unlike these works, we reveal the underlying anisotropic diffusion scheme associated with graph Laplacian regularization based on our convergence result. 

\subsubsection{Riemannian metric for inverse imaging}
In several works, {\it e.g.}, \cite{rosman12,sochen98,wetzler11}, continuous images are recovered based on Riemannian manifolds with metrics derived from local image contents. These works apply numerical methods to approximate the Beltrami flow---generalization of heat flow on Riemannian manifold, where it is necessary to explicitly estimate the Riemannian metric.
Although we also derive an optimal metric with similar functional form, our work is essentially a \emph{graph-based method}---we never explicitly compute the optimal metric. To build the optimal graph Laplacian for denoising, our work only needs the discrete feature functions.
In addition, our derived optimal metric utilizes \emph{non-local} information to denoise the images effectively.

%% file: interp_xlx.tex


We first construct an underlying graph $\mathcal{G}$ that supports a graph-signal $\mathbf{u}$ on top. 
We then demonstrate the convergence of the graph Laplacian regularizer $S_\mathcal{G}({\bf{u}}) = \bf{u}^{\rm{T}} \mathbf{L} \, \mathbf{u}$ to the anisotropic Dirichlet energy $S_{\Omega}$ \cite{flucher99}---a quadratic functional in the continuous domain. 
We then analyze in detail the functional $S_{\Omega}$ to understand its discrete counterpart $S_\mathcal{G}$.

\subsection{Graph Construction from Exemplar Functions}
\label{ssec:graph}

\begin{figure}[!t]
\centering
    \includegraphics[height=75pt]{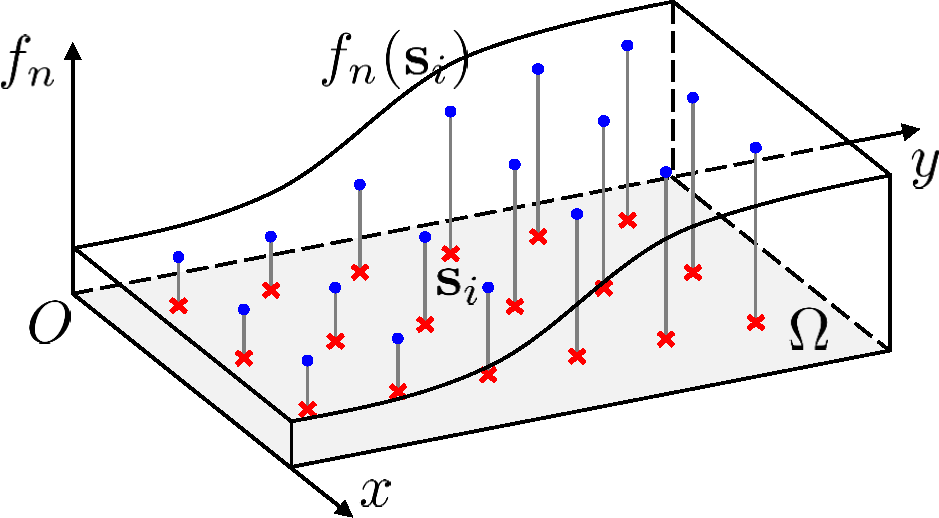}
\caption{Sampling the exemplar function $f_n$ at pixel locations in domain $\Omega$.}
\label{fig:samp}
\end{figure}

To facilitate understanding, we describe the construction of our discrete graph ${\cal{G}}$ and define corresponding continuous quantities in parallel. 
We first define $\Omega$, a bounded region in $\mathbb{R}^2$, as the \emph{domain} on which a continuous image (or image patch) lives. 
In practice, $\Omega$ takes a rectangular shape; see Fig.\,\ref{fig:samp} for an illustration. 
Denote by $\Gamma=\{{\bf{s}}_i=(x_i,y_i)~|~{\bf{s}}_i\in\Omega, 1\le i\le M\}$ a set of $M$ \textit{random coordinates} uniformly distributed on $\Omega$ (\textit{e.g.}, red crosses in Fig.\,\ref{fig:samp}). Since pixel coordinates are uniformly distributed on an image, we interpret the collection of pixel coordinates as one possible set $\Gamma$.

For any location ${\bf s}=(x,y)\in\Omega$, we denote by $f_n({\bf{s}}):\Omega\mapsto\mathbb{R}$, $1\le n \le N$, a set of $N$ continuous functions defined on $\Omega$; we will call them \textit{exemplar functions}.
These functions, which can be freely chosen by the users, are critical in determining graph connections and edge weights.
One obvious choice for the exemplar functions is the estimates/observations of the desired ground-truth signal. 
For example, in image denoising where a noisy patch is given, the $f_n$'s can be the noisy patch itself and another $K-1$ non-local similar patches due to self-similarity of natural images. 
Hence in this case, there are $N=K$ exemplar functions. 
However, this selection turns out to be sub-optimal. 
In this work, we will develop a methodology to choose $f_n$'s optimally in Section~\ref{sec:optim_xlx}.

By sampling the exemplar functions $\{f_n\}_{n=1}^N$ at coordinates $\Gamma$, $N$ {\it{discrete}} exemplar functions of length $M$ are obtained:
\begin{equation}\label{eq:fD}
\mathbf{f}_n=[f_n({\bf{s}}_1)\ f_n({\bf{s}}_2)\ \dots\ f_n({\bf{s}}_M)]^\mathrm{T},
\end{equation}
where $1\le n \le N$. Fig.\,\ref{fig:samp} illustrates the sampling process of an exemplar  function $f_n$---a simple ramp in $\Omega$. The blue dots are samples of $f_n$ and collectively form the vector $\mathbf{f}_n$.

For each pixel location ${\bf{s}}_i\in\Gamma$, we construct a length $N$ vector $\mathbf{v}_i$ ($1\le i \le M$) using the previously defined $\mathbf{f}_n$,
\begin{equation}\label{eq:sample}
\mathbf{v}_i=[\mathbf{f}_1(i)\ \mathbf{f}_2(i)\ \dots\ \mathbf{f}_N(i)]^{\rm{T}},
\end{equation}
where we denote the $i$-th entry of $\mathbf{f}_n$ as $\mathbf{f}_n(i)$, so $\mathbf{f}_n(i)=f_n({\bf s}_i)$. With vectors $\{\mathbf{v}_i\}_{i=1}^M$, we build a weighted neighborhood graph $\mathcal{G}$ with $M$ vertices, where each pixel location ${\bf{s}}_i\in\Gamma$ is represented by a vertex $V_i\in{\cal V}$. The weight $w_{ij}$ between two different vertices $V_i$ and $V_j$ is computed as
\begin{equation}\label{eq:weight}
{w_{ij}} = {{({\rho_i}{\rho_j})}^{-\gamma} }{\psi (d_{ij})}.
\end{equation}
The weighting kernel $\psi(\bm\cdot)$ is a thresholded Gaussian function
\begin{equation}\label{eq:kernel}
\psi (d_{ij}) = \left\{ {\begin{array}{*{20}{l}}
{\exp \left( { - \frac{\displaystyle{{d_{ij}^2}}}{\displaystyle{2\epsilon{^2}}}} \right)} & {\mbox{if} ~~ \left| d_{ij} \right| \le r,}\\
0&{{\rm{otherwise,}}}
\end{array}} \right.
\end{equation}
and
\begin{equation}\label{eq:graph_rest}
d_{ij}^2={{\left\| {{{\bf{v}}_i} - {{\bf{v}}_j}} \right\|}_2^2},\hspace{5pt}{\rho_i} = \sum\nolimits_{j = 1}^M {\psi (d_{ij})},
\end{equation}
where $d_{ij}^2$ measures the Euclidean distance between two vertices in the space defined by the exemplar functions, and the constant $\epsilon$ controls the sensitivity of the graph weights to the distances. 
The term ${{({\rho_i}{\rho_j})}^{-\gamma} }$ re-normalizes the graph weight with the normalization parameter $\gamma$, where ${\rho_i}$ is the degree of vertex $V_i$ before normalization. 
As mentioned in Section\,\ref{ssec:related_other}, we show our convergence result based on \cite{hein06}. 
Hence, similar to \cite{hein06}, we also introduce the normalization term ${{({\rho_i}{\rho_j})}^{-\gamma} }$ when defining edge weights. To be shown in Section\,\ref{ssec:pws}, it is particularly useful since different extents of normalization result in different denoising effects.
Under these settings, $\mathcal{G}$ is an \emph{$r$-neighborhood graph}, \textit{i.e.}, there is no edge connecting two vertices with a distance greater than $r$. Here $r=\epsilon C_r$, and $C_r$ is a constant.
We note that graphs employed in many recent works ({\it{e.g.}}, \cite{kheradmand13,hu14,liu14} and \cite{wan14}) are special cases of our more generally defined graph $\mathcal{G}$.

With the constructed graph, we can now define the \emph{adjacency matrix} ${\bf A}\in\mathbb{R}^{M\times M}$, which is a symmetric matrix with $w_{ij}$ as its $(i,j)$-th entry. 
The \emph{degree matrix} $\bf{D}$ of graph $\mathcal{G}$ is a diagonal matrix with its $i$-th diagonal entry computed as $\sum\nolimits_{j = 1}^M {{w_{ij}}}$. Then the \emph{unnormalized graph Laplacian} \cite{shuman13}---the most basic type of graph Laplacian---is given by
\begin{equation}\label{eq:graph_laplacian}
{\bf L}={\bf D}-{\bf A}.
\end{equation}
${\bf L}$ has $0$ as its smallest eigenvalue and a constant vector as the corresponding eigenvector; it is symmetric and positive semi-definite \cite{shuman13}.
\begin{figure}[!t]
\centering
    \includegraphics[height=55pt]{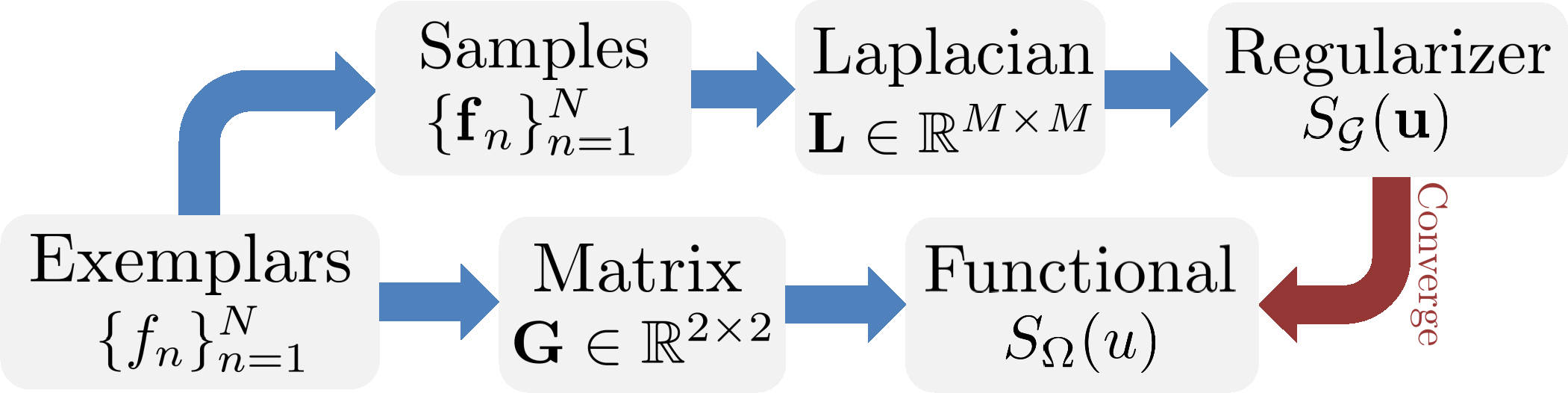}
\caption{Relationships among key quantities, where a blue arrow pointing from block $A$ to block $B$ means $B$ is derived from $A$.}
\label{fig:rela}
\end{figure}

\subsection{Graph Laplacian Regularizer and its Convergence}
\label{ssec:conv}

We now formally define the graph Laplacian regularizer and show its convergence to a functional for 2D images in the continuous domain. 
Denote by $u(x,y):\Omega\mapsto\mathbb{R}$ a smooth\footnote{``smooth" here means a function with derivatives of all orders.} \textit{candidate function} defined in domain $\Omega$.
Sampling $u$ at positions of $\Gamma$ leads to its discretized version, $\mathbf{u}=[u({\bf{s}}_1)\ u({\bf{s}}_2)\ \dots\ u({\bf{s}}_M)]^\mathrm{T}$. 
Using $\mathbf{L}$, the graph Laplacian regularizer for $\mathbf{u}$ can now be written as $S_{\mathcal{G}}(\mathbf{u}) = {\mathbf{u}}^{\rm{T}}{\bf{L}} \, \mathbf{u}$. 
Recall that $\cal{E}$ is the set of edges, it can be shown that
\begin{equation}
\label{fig:xlx_explain}
{S_\mathcal{G}}({\bf{u}}) = {{\bf{u}}^{\rm{T}}}{\bf{Lu}} = \sum\limits_{(i,j) \in \mathcal{E}} {{w_{ij}}{{\left( {{u({\bf s}_i)} - {u({\bf s}_j)}} \right)}}^2}.
\end{equation}
%
$S_{\mathcal{G}}(\mathbf{u})$ is small when signal $\bf{u}$ has similar intensities between vertices connected by edges with large weights.
Hence minimizing the graph Laplacian regularizer imposes smoothness on $\bf{u}$ with respect to the graph $\cal{G}$ \cite{shuman13}.

The continuous counterpart of regularizer $S_{\mathcal{G}}(\mathbf{u})$ is given by a functional $S_{\Omega}(u)$ for function $u$ defined in domain $\Omega$,
\begin{equation}\label{eq:confunc}
{S_\Omega }(u) = \int_\Omega {{{\nabla u}^{\rm{T}}}{{\bf{G}}^{ - 1}}\nabla u{{\left( {\sqrt {\det {\bf{G}}} } \right)}^{2\gamma  - 1}}}d{\bf{s}},
\end{equation}
where $\nabla u = [{\partial _x}u\hspace{5pt}{\partial _y}u]^{\rm{T}}$ is the gradient of continuous function $u$, and ${\bf s}=(x,y)$ is a location in $\Omega$. 
Recall that $\gamma$ is the normalization parameter introduced in \eqref{eq:weight}. 
$S_{\Omega}(u)$ is also called the \emph{anisotropic Dirichlet energy} in the literature \cite{alliez07,flucher99}, and $\mathbf{G}$ is a 2$\times$2 matrix:
\begin{equation}\label{eq:metric}
\begin{split}
{\bf{G}}\hspace{-1pt}=\hspace{-1pt}\sum_{n = 1}^N{\left[\hspace{-5pt}{\begin{array}{*{20}{c}}{{{\left( {{\partial _x}{f_n}} \right)}^2}}&\hspace{-8pt}{{\partial _x}{f_n}\bm\cdot{\partial _y}{f_n}}\\
{{\partial _x}{f_n}\bm\cdot{\partial _y}f}&\hspace{-8pt}{{{\left( {{\partial _y}{f_n}} \right)}^2}}\end{array}}\hspace{-5pt}\right]}\hspace{-3pt}=\hspace{-3pt}\sum\nolimits_{n = 1}^N\hspace{-5pt}{\nabla {f_n}\hspace{-1pt}\bm\cdot\hspace{-1pt}{{\nabla {f_n^{\rm{T}}}}}}.
\end{split}
\end{equation}
$\mathbf{G}:\Omega\mapsto\mathbb{R}^{2\times2}$ is a matrix-valued function of location ${\bf{s}}\in\Omega$.
It can be viewed as the \textit{structure tensor}~\cite{knutsson11} of the gradients $\{\nabla f_n\}_{n=1}^N$.
The computation of $\mathbf{G}$ is also similar to that of the covariance matrix used in the \emph{steering kernel} \cite{takeda07}, though our matrix $\mathbf{G}$ is computed from a more general set of exemplar functions.
Note that the exemplar functions $\{f_n\}_{n=1}^N$ exactly determine the functional $S_\Omega$ and the graph Laplacian regularizer $S_{\mathcal{G}}$.
\begin{figure}[!t]
\centering
    \subfigure[]{\includegraphics[height=80pt]{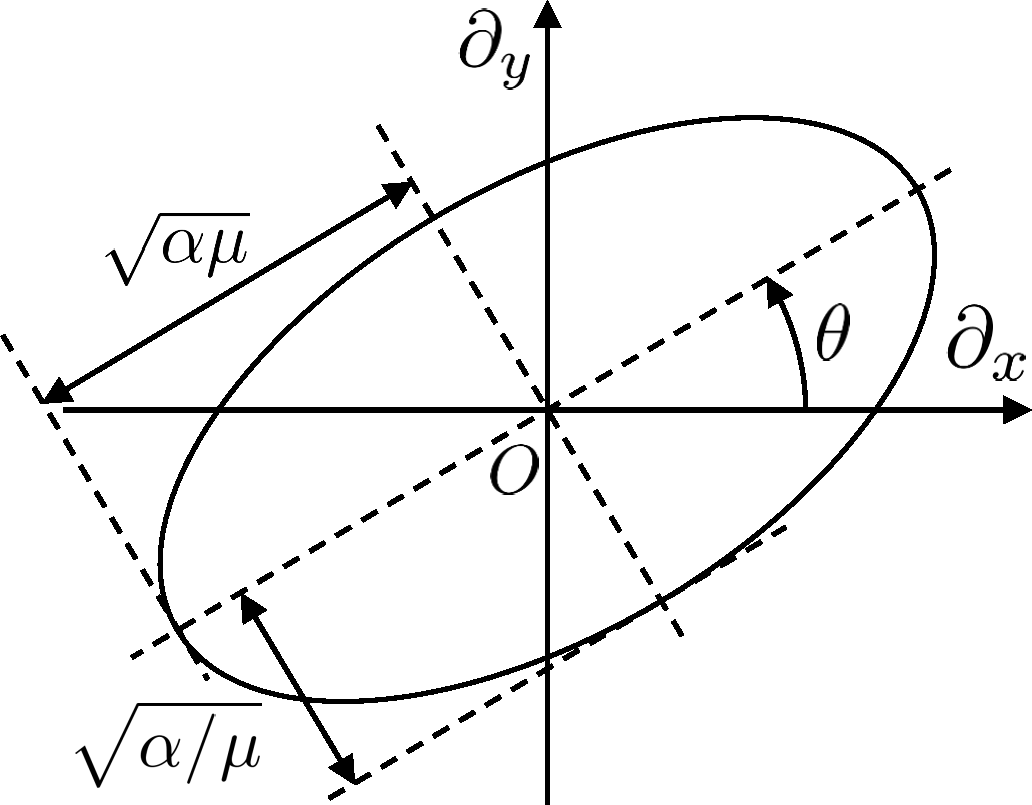}\label{fig:metric_a}}\hspace{5pt}
    \subfigure[]{\includegraphics[height=80pt]{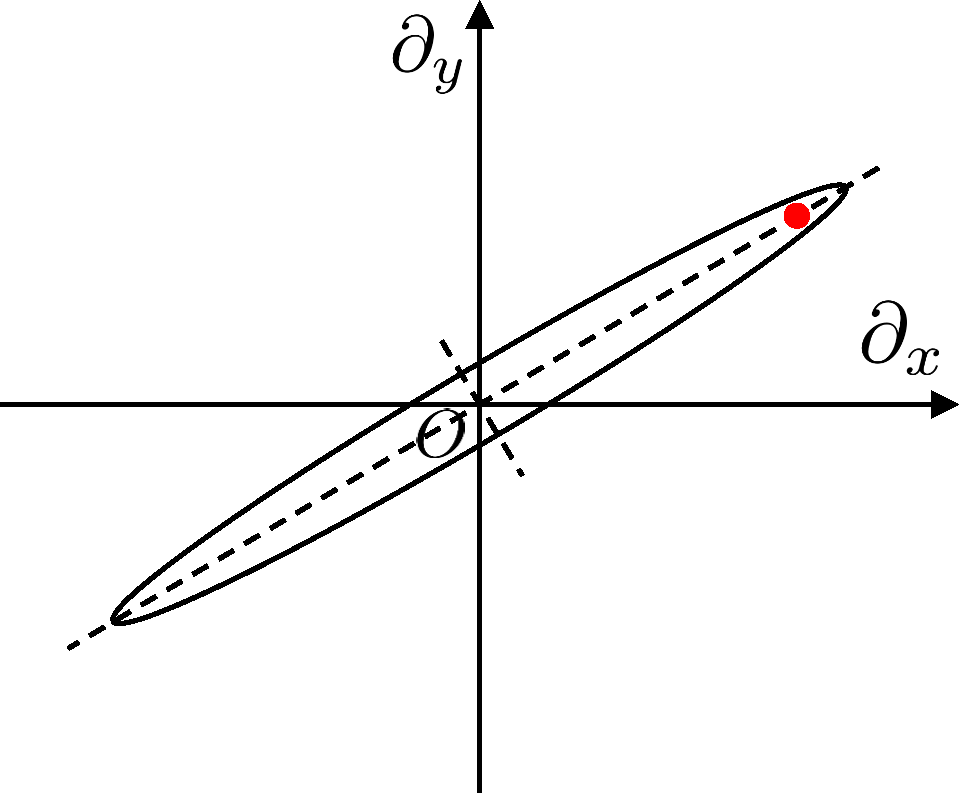}\label{fig:metric_b}}
\caption{Unit-distance ellipses of (a) a general metric space $\bf{G}$; and (b) an ideal metric space ${\bf G}_I$ in the gradient coordinates, where the red dot marks the ground-truth gradient.}
\label{fig:metric}
\end{figure}

We can now declare the following theorem:
\begin{theorem}[Convergence of $S_{\mathcal{G}}$]\label{thm:conv}
Under mild conditions for $\epsilon$, functions $\{f_n\}_{n=1}^N$ and $u$ as stated in Appendix\,\ref{app:conv_proof},
\begin{equation}\label{eq:conv}
\mathop {\lim }\limits_{\genfrac{}{}{0pt}{}{\scriptstyle M \to \infty}{\scriptstyle \epsilon \to 0} }{S_{\mathcal{G}}}(\mathbf{u}) \sim {S_{\Omega}}(u),
\end{equation}
where ``$\sim$'' means there exists a constant depending on $\Omega$, $C_r$, and $\gamma$, such that the equality holds.
\end{theorem}
\noindent In other words, as the number of samples $M$ increases and the neighborhood size $r=\epsilon C_r$ shrinks, the graph Laplacian regularizer $S_{\mathcal{G}}$ approaches the anisotropic Dirichlet energy $S_\Omega$. 
To prove Theorem\;\ref{thm:conv}, we regard the graph $\cal{G}$ as a discrete approximation of a Riemannian manifold $\cal{M}$, where $\cal{M}$ is a 2D manifold embedded in $\mathbb{R}^N$ with coordinates $\left(f_1,f_2,\ldots,f_N\right)$ in $\mathbb{R}^N$.
Then, the above theorem can be proven based on the result in \cite{hein06}. 
We provide the proof in Appendix\,\ref{app:conv_proof}.\footnote{For the sake of intuitive presentation, we weaken the uniform convergence of the proof to point-wise convergence in \eqref{eq:conv}.}
The relationships of several key quantities in our analysis are summarized in Fig.\,\ref{fig:rela}.

\subsection{Metric Space in the Continuous Domain}
\label{ssec:analysis}
\begin{figure}[!t]
\centering
    \subfigure[]{\includegraphics[height=55pt]{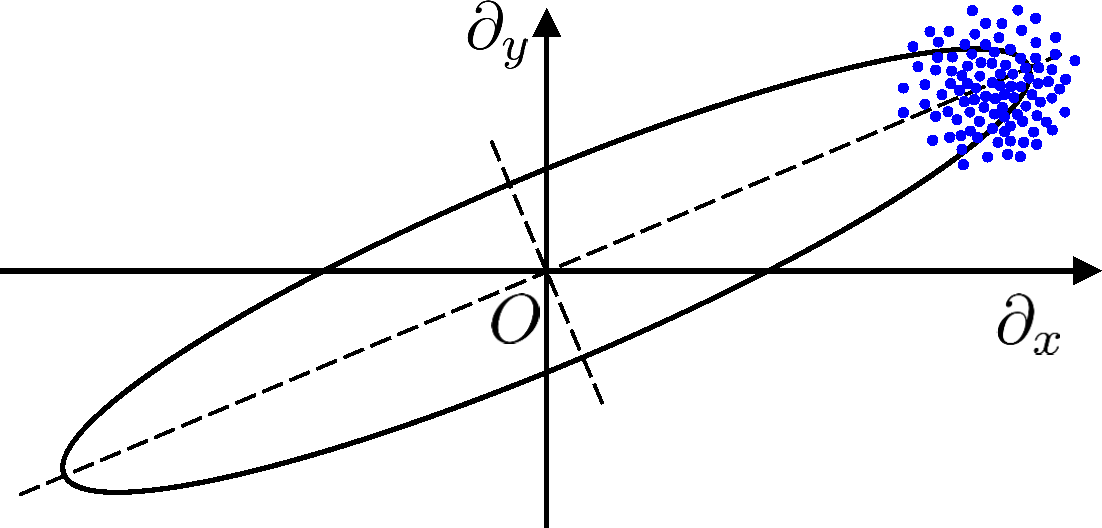}\label{fig:pca_a}}\hspace{5pt}
    \subfigure[]{\includegraphics[height=55pt]{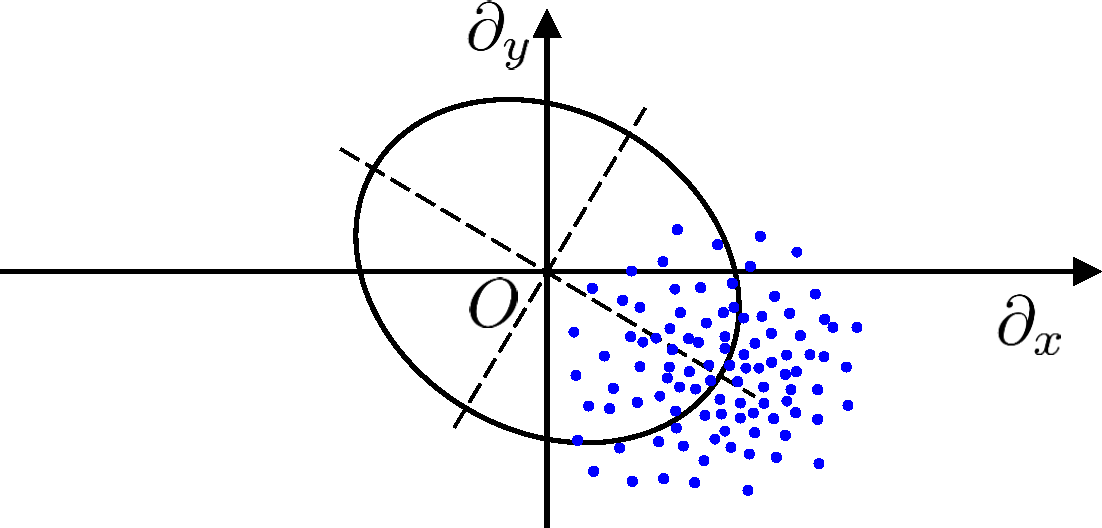}\label{fig:pca_b}}
\caption{Different distributions of points $\{\nabla f_n\}_{n=1}^N$ establish different metric spaces. (a) A densely distributed set of gradients results in a more skewed metric space. (b) A scattered set of gradients gives a less skewed metric space.
The unit-distance ellipses in both diagrams are shrunk for better visualization.
}
\label{fig:pca}
\end{figure}

The convergence of the graph Laplacian regularizer $S_{\mathcal{G}}$ to the anisotropic Dirichlet energy $S_\Omega$ allows us to understand the mechanisms of $S_{\mathcal{G}}$ by analyzing $S_\Omega$.
From \eqref{eq:confunc}, the quadratic term ${{{\nabla u}^{\rm{T}}}{\bf G}^{-1}\nabla u}$ measures the length of gradient $\nabla u$ in a \emph{metric space} determined by matrix ${\bf{G}}$;  it is also the \emph{Mahalanobis distance} between the point $\nabla u$ and the distribution of the points $\{\nabla f_n\}_{n=1}^N$ \cite{mahalanobis36}.
In the following, we slightly abuse the notation and call $\bf{G}$ the \emph{metric space}. Similar to the treatment for the steering kernel\cite{takeda07}, we perform eigen-decomposition to ${\bf G}$ to analyse $S_\Omega$:
\begin{align}
{\bf{G}} &= \alpha {\bf{U\Lambda }}{{\bf{U}}^{\rm{T}}},\label{eq:G_eig}\\
{\bf{U}} &=\begin{bmatrix} 
\cos \theta & -\sin \theta\\
\sin \theta & \cos \theta 
\end{bmatrix},\hspace*{5pt}
{\bf{\Lambda}} =\begin{bmatrix} 
\mu & 0 \\
0 & \mu ^{ - 1}
\end{bmatrix},\label{eq:G_eig_2}
\end{align}
where $\mu\ge1$, $\theta\in[0,\pi)$ and $\alpha>0$. 
One can verify that the \emph{unit-distance ellipse}---the set of points having distance $1$ from the origin \cite{ono94}---of metric space $\bf{G}$ is an ellipse with semi-major axis $\sqrt{\alpha\mu}$ and semi-minor axis $\sqrt {\alpha/\mu }$.
Fig.\,\ref{fig:metric_a} shows the unit-distance ellipse of metric space ${\bf G}$ in the gradient coordinates.
From \eqref{eq:G_eig} and \eqref{eq:G_eig_2}, metric space $\bf{G}$ is uniquely determined by parameters $\{\mu,\theta,\alpha\}$ illustrated as follows:
\begin{enumerate}[(i)]
\item{\emph{Skewness} $\mu$: a bigger $\mu$ results in a more skewed metric and a more elongated unit-distance ellipse.}
\item{\emph{Major direction} $\theta$: along direction $\theta$ the metric norm increases the slowest. We call its perpendicular direction the \emph{minor direction}, along which the metric norm increases the fastest.}
\item{\emph{Scaling parameter} $\alpha$: the value of $\alpha$ describes how fast the metric norm increases; a smaller $\alpha$ means the metric increases faster and the unit-distance ellipse is smaller.}
\end{enumerate}
For the same length $|\nabla u|$, we see that $\nabla u^{\rm{T}} {\mathbf{G}}^{-1} \nabla u$ computes to different values for $\nabla u$ with different directions.
The Euclidean space is a special case of $\bf{G}$ by letting $\alpha=\mu=1$, whose unit-distance ellipse is a unit circle.

In addition, establishing the metric space $\bf{G}$ is similar to using \emph{principal component analysis} (PCA) to analyze the set of $N$ points $\{\nabla f_n\}_{n=1}^{N}$. 
Intuitively, the skewness of a metric (``elongation'' of the unit-distance ellipse)  reflects the ``concentration'' of $\{\nabla f_n\}_{n=1}^N$, and the major direction aligns with the ``center'' of $\{\nabla f_n\}_{n=1}^N$.
The size of the unit-distance ellipse reflects the magnitudes of $\{\nabla f_n\}_{n=1}^N$. 
Fig.\,\ref{fig:pca} illustrates the impact of different point sets $\{\nabla f_n\}_{n=1}^N$ on the metric spaces, where the blue dots are gradients $\{\nabla f_n\}_{n=1}^N$, and the ellipses are the unit-distance ellipses. 
We see that a densely distributed set of gradients results in a more skewed metric space (Fig.\,\ref{fig:pca_a}), while a scattered set of gradients leads to a less skewed metric space (Fig.\,\ref{fig:pca_b}). 

\subsection{Continuous Functional as the Regularizer}
\label{ssec:reg_metric}
According to the convergence result \eqref{eq:conv}, using the graph Laplacian regularizer in the discrete domain corresponds to using the functional $S_\Omega$ as a regularizer in the continuous domain.
From the expression of $S_\Omega$ \eqref{eq:confunc}, this further boils down to using the metric norm $\nabla {u^{\rm{T}}}{{\bf G}^{ - 1}}\nabla u$ as a regularizer on a point-by-point basis throughout the image domain $\Omega$.
\begin{figure}[!t]
\centering
    \subfigure[]{\includegraphics[height=65pt]{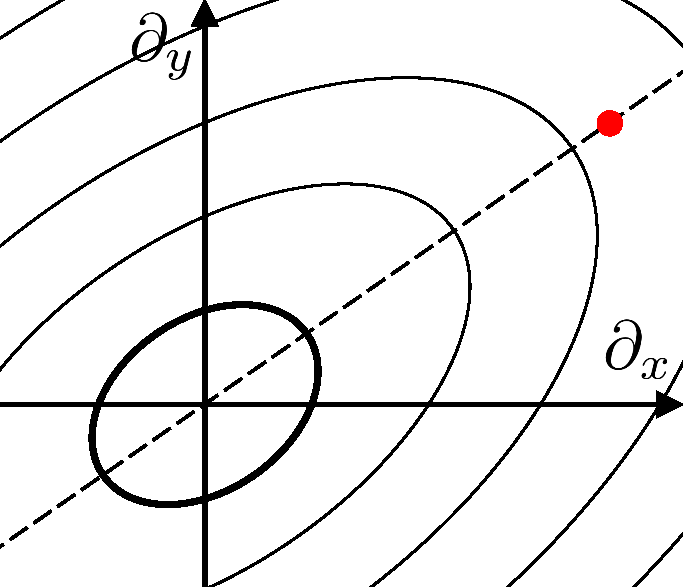}\label{fig:reg_a}}\hspace{5pt}
    \subfigure[]{\includegraphics[height=65pt]{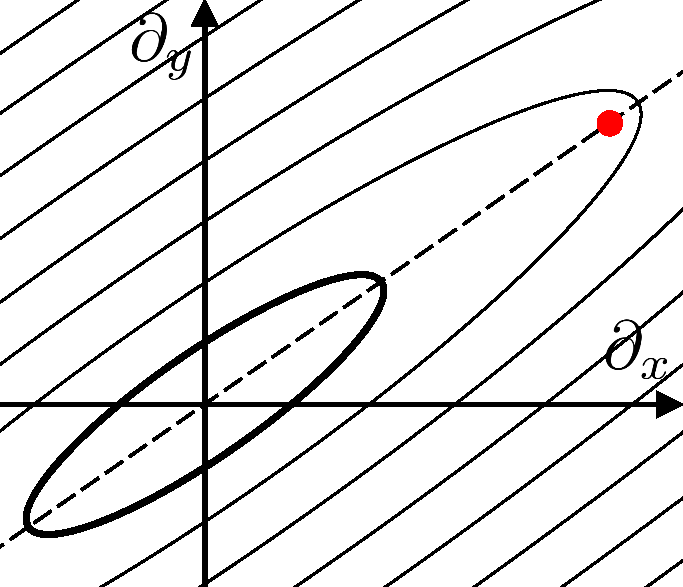}\label{fig:reg_b}}\hspace{5pt}
    \subfigure[]{\includegraphics[height=65pt]{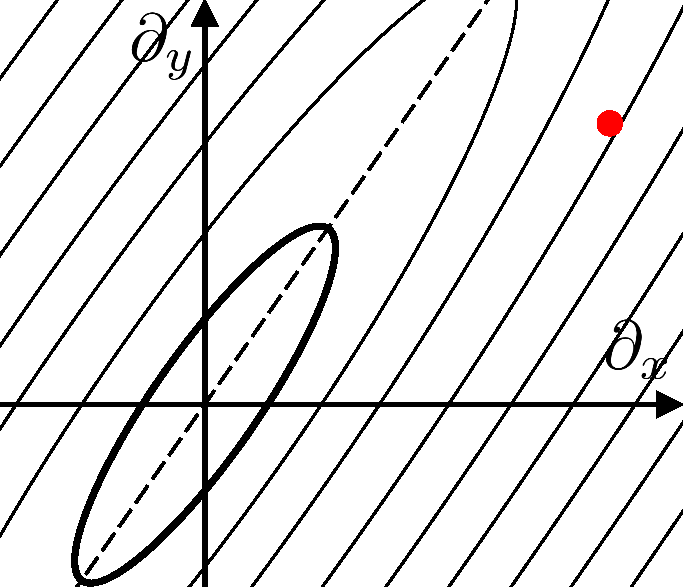}\label{fig:reg_c}}
\caption{Different scenarios of using the metric norm as a ``pointwise'' regularizer. The red dots mark the ground-truth gradient.}
\label{fig:reg_metric}
\end{figure}

Fig.\;\ref{fig:reg_metric} shows different scenarios of applying the metric norm $\nabla {u^{\rm{T}}}{{\bf G}^{ - 1}}\nabla u$ as a ``point-wise'' regularizer.
Denote by $\mathbf{g}$ the ground-truth gradient of the original image, which is marked with a red dot in each plot. 
We also draw the contour lines of the metric spaces, where the most inner (bold) ones are the unit-distance ellipses. 
We see that, though both metric spaces in Fig.~\ref{fig:reg_a} and Fig.~\ref{fig:reg_b} have major directions aligned with ${\bf g}$, Fig.~\ref{fig:reg_b} is more skewed, and hence more \emph{discriminant}, {\it{i.e.}}, a small Euclidean distance away from $\mathbf{g}$ along the minor direction of ${\bf G}$ results in a large metric distance. 
It is \textit{desirable} for a regularizer to distinguish between good image patch candidates (close to ground-truth) and bad candidates (far from ground-truth). 
However, if the metric space is skewed but its major direction does not align with $\mathbf{g}$ (Fig.~\ref{fig:reg_c}), it is \textit{undesirable} because bad image patch candidates will have a smaller cost than good candidates.

As a result, for inverse imaging problems where $\bf{g}$ is unknown, one should design a \emph{robust} metric space $\mathbf{G}$ based on an initial estimate of {\bf g}, such that:
\begin{enumerate}[(i)]
\item{$\bf{G}$ has a major direction aligned with the estimate, {\it i.e.}, it is discriminant with respect to the estimate;}
\item{The metric space $\bf{G}$ is discriminant only to the extent that the estimate is reliable.}
\end{enumerate}
The notion of metric space allows us to understand what signals are being discriminated and to what extent on a point-by-point basis, which explains the mechanisms of the graph Laplacian regularizer in the continuous domain.

Finally, we note that from the definition of $S_\Omega$ \eqref{eq:confunc}, the original scaling parameter $\alpha$ is re-normalized as $\alpha\bm\cdot{{\left( {\sqrt {\det {\bf{G}}} } \right)}^{1-2\gamma}}=\alpha^{2(1-\gamma)}$ by the normalization parameter $\gamma$ (note that $\det {\bf{G}} = \alpha^2$ from \eqref{eq:G_eig} and \eqref{eq:G_eig_2}). Interestingly, $\gamma$ also re-normalizes the graph weights \eqref{eq:weight} in the discrete domain. Under the context of \emph{anisotropic diffusion}~\cite{weickert98}, Section\,\ref{sec:diffu_xlx} will provide a thorough analysis of the effects of choosing different $\gamma$'s.

%% file: optim_xlx.tex
Equipped with the analysis of $S_\Omega$, we now derive the \emph{optimal} graph Laplacian regularizer for image denoising via a patch-based non-local approach \cite{hu13,pang14,buades05}. 
Because we denoise an input image on a patch-by-patch basis, the domain $\Omega$ is a square region accommodating continuous image patches in our method.
We first establish an ideal metric space ${\bf{G}}_I$ given the ground-truth gradient {\bf{g}}. 
Next, we introduce a noise model (independent and identically distributed (i.i.d.) additive white Gaussian noise (AWGN)) in the \emph{gradient} domain. 
With a set of noisy but similar non-local gradient observations, we then derive the \emph{optimal} metric space ${\bf{G}}^\star$ in the {\emph{minimum mean square error}} (MMSE) sense. From ${\bf{G}}^\star$, we then derive the corresponding optimal exemplar functions $\{f_n^\star\}_{n=1}^N$ according to the metric space definition \eqref{eq:metric}. Their discrete counterparts, $\{{\bf{f}}_n^\star\}_{n=1}^N$, are then used to compute the optimal graph Laplacian $\mathbf{L}^\star$ for graph Laplacian regularization in \eqref{eq:prob_discrete}.

\subsection{Ideal Metric Space}\label{ssec:ideal_space}
We first establish the ideal metric space $\mathbf{G}_I$ when the ground-truth gradient $\mathbf{g}({\bf{s}})$ at location $\bf{s}$, ${\bf{s}}\in\Omega$, is known:
\begin{equation}\label{eq:G_atom}
{{\bf{G}}_I}({\bf{g}}) = {\bf{g}}{{\bf{g}}^{\rm{T}}} + \beta {\bf{I}},
\end{equation}
where $\beta$ is a small positive constant. 
The quantity $\beta {\bf{I}}$ is included in \eqref{eq:G_atom} to ensure that the metric space ${\bf G}_I$ is well-defined---{\it i.e.}, ${\bf G}_I$ is invertible, so $\nabla {u^{\rm{T}}}{{\bf G}_I^{ - 1}}\nabla u$ is computable. 
In fact, when $\mathbf{g}(\mathbf{s}) = \mathbf{0}$, {\it e.g.}, in flat regions, ${\bf{G}}_I=\beta{\bf I}$, corresponding to a scaled Euclidean space. By performing eigen-decomposition, as similarly done in Section \ref{ssec:analysis}, we can see that ${\bf G}_I$ has a major direction aligned with ${\bf g}$. Moreover, the skewness and scaling parameters of ${{\bf{G}}_I}({\bf{g}})$---denoted by $\mu_I$ and $\alpha_I$, respectively---are given by
\begin{equation}\label{eq:metric_parm}
\mu _I^2 = 1 + \left\| {\bf{g}} \right\|_2^2/\beta,\hspace{10pt}\alpha _I^2 = \beta \left\| {\bf{g}} \right\|_2^2 + {\beta ^2},
\end{equation}
according to \eqref{eq:G_eig} and \eqref{eq:G_eig_2}. Hence, the skewness of ${\bf{G}}_I$ can be adjusted using $\beta$, where a smaller $\beta$ means a more skewed metric space. Since the ground truth ${\bf g}$ is known, it is desirable to have a very skewed metric space---$\beta$ should be very small. For illustration, Fig.\,\ref{fig:metric_b} shows an ideal metric space with an elongated unit-distance ellipse.

%
%

%
%
%
\subsection{Noise Modeling of the Patch Gradients}
\label{ssec:gradnoi}

Like previous self-similarity assumptions in \cite{buades05,dabov07}, etc., we also assume that similar pixel patches recur throughout an image. 
Specifically, given a $\sqrt{M}\times\sqrt{M}$ noisy \textit{target patch} ${\bf{z}}_0\in\mathbb{R}^M$, we assume that there exists a set of $K-1$ non-local patches in the noisy image that are similar to ${\bf{z}}_0$ in terms of \emph{gradients}.
Together with ${\bf{z}}_0$, the $K$ patches $\{{\bf{z}}_k\}_{k=0}^{K-1}$ are collectively called a \emph{cluster} in the sequel.
We denote the continuous counterpart of patch ${\bf{z}}_k$ as $z_k({\bf{s}}):\Omega\mapsto\mathbb{R}$, $0\le k\le K-1$, and represent $\nabla z_k({\bf{s}})$---the gradient of $z_k$ at location ${\bf{s}}\in\Omega$---as ${\bf{g}}_k({\bf{s}})$. 
The variable $\bf{s}$ is omitted hereafter for simplicity.

As analyzed in \cite{zhang15}, AWGN in the gradient domain is approximately equivalent to AWGN in the spatial domain. 
For simplicity,  herein we introduce AWGN in the gradient domain, as similarly done in \cite{zhang15} and \cite{karaccali04}.
With the cluster $\{{\bf{z}}_k\}_{k=0}^{K-1}$, we model the noisy gradients $\{{\bf{g}}_k\}_{k=0}^{K-1}$ at a location ${\bf{s}}\in\Omega$ as
\begin{equation}\label{eq:noise}
{\bf{g}}_k={\bf{g}}+{\bf{e}}_k,\hspace{3pt} 0\le k\le K-1,
\end{equation}
where ${\bf{g}}$ is the ground-truth (noiseless) gradient at $\bf{s}$ to be recovered. 
$\{{\bf{e}}_k\}_{k=0}^{K-1}$ are i.i.d. noise terms in the gradient domain, which follow a 2D Gaussian distribution with zero-mean and covariance matrix ${\sigma_g^2}{\bf{I}}$ (${\bf{I}}$ is the $2\times 2$ identity matrix).
So the probability density function (PDF) of ${\bf{g}}_k$ given ${\bf{g}}$ is
\begin{equation}\label{eq:gra_dist}
Pr( {{{\bf{g}}_k}\given[\big]{\bf{g}}} ) = \frac{1}{{2\pi {\sigma_g ^2}}}\exp \left( { - \frac{1}{{2{\sigma_g ^2}}}\left\| {{\bf{g}} - {{\bf{g}}_k}} \right\|_2^2} \right).
\end{equation}
We assume that $\sigma_g^2$ is constant over $\Omega$, though it can be different for different clusters.
We will introduce a procedure in Section\;\ref{sec:alg} to identify similar patches $\{{\bf{z}}_k\}_{k=1}^{K-1}$ in the image given ${\bf{z}}_0$, and to estimate a proper $\sigma_g^2$ for each cluster given that the image is corrupted by AWGN in the pixel domain.

\subsection{Seeking the Optimal Metric Space}\label{ssec:opt_met}
Given the noisy gradients $\{{\bf{g}}_k\}_{k=0}^{K-1}$, we seek the optimal metric space $\mathbf{G}^{\star}$ in the MMSE sense. We consider the following minimization problem:
\begin{equation}\label{eq:mmse}
{{\bf{G}}^ \star }\hspace{-2pt}=\hspace{-1pt}\mathop {\arg \min }\limits_{\bf{G}}\hspace{-2pt} \int_{\mathbb{R}^2}\hspace{-3pt}{\left\| {{\bf{G}}\hspace{-1pt}-\hspace{-1pt}{{\bf{G}}_I}({\bf{g}})} \right\|_F^2\hspace{-1pt}\bm\cdot\hspace{-1pt}Pr\hspace{-2pt}\left({\bf{g}}\given[\Big]\hspace{-2pt}\left\{ {{{\bf{g}}_k}} \right\}_{k = 0}^{K - 1}\right)\hspace{-2pt}d{\bf{g}}},
\end{equation}
where the differences between metric spaces are measured by the \emph{Frobenius} norm; we choose the Frobenius norm for ease of optimization. 
By taking the derivative of the objective in \eqref{eq:mmse} with respect to $\bf{G}$ and setting it to zero, we obtain
\begin{equation}\label{eq:opt_org}
{\bf{G}}^{\star} = \int_{\mathbb{R}^2}  {{{\bf{G}}_I}({\bf{g}})\bm\cdot Pr \left({\bf{g}}\,\given[\Big]\left\{ {{{\bf{g}}_k}} \right\}_{k = 0}^{K - 1}\right)d{\bf{g}}}.
\end{equation}
This means that ${\bf{G}}^{\star}$ is the weighted average of ${{\bf{G}}_I}({\bf{g}})$ over the entire gradient domain $\mathbb{R}^2$.

Using Bayes' theorem, we replace the posterior probability of \eqref{eq:opt_org} with the product of the likelihood and prior:
\begin{equation}\label{eq:opt_der}
\begin{split}
&Pr\left( {{\bf{g}}\,\given[\Big]\left\{ {{{\bf{g}}_k}} \right\}_{k = 0}^{K - 1}} \right) \propto Pr({\bf{g}})\bm\cdot Pr\left( {\left\{ {{{\bf{g}}_k}} \right\}_{k = 0}^{K - 1}\given[\Big]\,{\bf{g}}} \right)\\
&\propto Pr({\bf{g}})\bm\cdot\prod\nolimits_{k = 0}^{K - 1} {Pr\left( {{{\bf{g}}_k}|{\bf{g}}} \right)}\\
&\propto \exp \hspace{-2pt}\left( \hspace{-2pt}{ - \frac{1}{{2\sigma _p^2}}\left\| {\bf{g}} \right\|_2^2} \hspace{-1pt}\right)\hspace{-2pt}\bm\cdot\hspace{-1pt}\exp \hspace{-2pt}\left(\hspace{-2pt} { - \frac{1}{{2\sigma _g^2}}\sum\nolimits_{k = 0}^{K - 1} \hspace{-2pt}{\left\| {{\bf{g}}\hspace{-1pt}-\hspace{-1pt}{{\bf{g}}_k}} \right\|_2^2} } \right),
\end{split}
\end{equation}
where we apply \eqref{eq:gra_dist} and assume that the prior $Pr({\bf{g}})$ follows a 2D zero-mean Gaussian with constant covariance $\sigma_p^2{\bf{I}}$. 
Here $\sigma_p^2$ is a constant over the whole noisy image. 
With \eqref{eq:opt_der}, one can derive that $Pr\left( {{\bf{g}}\,\given[\Big]\left\{ {{{\bf{g}}_k}} \right\}_{k = 0}^{K - 1}} \right)$ is also a 2D Gaussian:
\begin{equation}\label{eq:g_concret}
Pr\left( {{\bf{g}}\,\given[\Big]\left\{ {{{\bf{g}}_k}} \right\}_{k = 0}^{K - 1}} \right) = \frac{1}{{2\pi {{\widetilde{\sigma}} ^2}}}\exp \left( { - \frac{1}{{2{{\widetilde{\sigma}} ^2}}}\left\| {{\bf{g}} - {\widetilde{\bf{g}} }} \right\|_2^2} \right),
\end{equation}
where its mean is $\widetilde{{\bf{g}}}$ and covariance is ${\widetilde{\sigma}}^2{\bf{I}}$, expressed as
\begin{equation}\label{eq:sta_fin}
 {\widetilde{\bf{g}}} = \frac{1}{{K + {{\sigma _g^2} \mathord{\left/
 {\vphantom {{\sigma _g^2} {\sigma _p^2}}} \right.
 \kern-\nulldelimiterspace} {\sigma _p^2}}}}\sum\nolimits_{k = 0}^{K - 1} {{{\bf{g}}_k}},\hspace{5pt}
 {\widetilde{\sigma} ^2} = \frac{{\sigma _g^2}}{{K + {{\sigma _g^2} \mathord{\left/
 {\vphantom {{\sigma _g^2} {\sigma _p^2}}} \right.
 \kern-\nulldelimiterspace} {\sigma _p^2}}}}.
\end{equation}
Here ${\widetilde{\bf{g}}}$ averages the noisy gradients, and it can be viewed as an \emph{estimate} of the ground truth $\bf{g}$. 
$ {\widetilde{\sigma} ^2}$ is a constant in domain $\Omega$, and it decreases as the number of observations $K$ increases.
With \eqref{eq:G_atom} and \eqref{eq:g_concret}, the optimal metric space \eqref{eq:opt_org} can be derived in closed-form:
\begin{equation}\label{eq:G_closed}
{\bf{G}}^\star = {\widetilde{\bf{g}}}{\widetilde{\bf{g}}}^{\rm{T}} + \beta_{\cal G}{\bf{I}},
\end{equation}
where we denote the constant $\beta_{\mathcal{G}}={\widetilde{\sigma}}^2+\beta$.

From \eqref{eq:G_closed}, ${\bf{G}}^\star$ has a major direction aligned with the estimate ${\widetilde{\bf{g}}}$. It has an intuitive interpretation: when the noise variance ${\widetilde{\sigma} ^2}$ is small, the first term dominates and the metric space is \emph{skewed and discriminant}; when ${\widetilde{\sigma} ^2}$ is large, \textit{i.e.}, the estimated gradient ${\widetilde{\bf{g}}}$ is unreliable, the second term dominates and the metric space is not skewed and is close to a \emph{non-discriminant Euclidean space}. 
Such properties of the optimal metric space ${\bf{G}}^\star$ are consistent with the analysis of designing robust metric spaces discussed in Section \ref{ssec:reg_metric}.
\subsection{From Metric Space to Graph Laplacian}\label{ssec:met2lap}
Continuous-domain notions, {\it e.g.}, the metric space and the average gradient $\widetilde{\bf g}$, are very useful for analysis. 
Nevertheless, when operating on discrete images, we need discrete exemplar functions $\{{\bf{f}}_n\}_{n=1}^{N}$ to compute the graph weights and obtain the graph Laplacian $\bf{L}$, as discussed in Section \ref{ssec:graph}.
Given \eqref{eq:metric}, which relates exemplar functions to a metric space, there exists a natural assignment of $N=3$ exemplar functions leading to the optimal metric space \eqref{eq:G_closed}.
Let
\begin{equation}\label{eq:fea_spat_con}
{f_1^\star}(x,y) = \sqrt {\beta_{\mathcal{G}}} \bm\cdot x,\hspace{5pt}{f_2^\star}(x,y) = \sqrt {\beta_{\mathcal{G}}} \bm\cdot y.
\end{equation}
According to \eqref{eq:metric}, $f_1^\star(x,y)$ and $f_2^\star(x,y)$ correspond to the term $\beta_{\cal G}{\bf{I}}$ in \eqref{eq:G_closed}. In the discrete domain,
\begin{equation}\label{eq:fea_spat}
{\bf{f}}_1^\star(i) = \sqrt {\beta_{\mathcal{G}}} \bm\cdot{x_i},\hspace{5pt}{\bf{f}}_2^\star(i) = \sqrt {\beta_{\mathcal{G}}} \bm\cdot{y_i}.
\end{equation}
Recall that $(x_i,y_i)$ are the coordinates of pixel $i$. 
Further, let
\begin{equation}\label{eq:fea_inten}
{f_3^\star}(x,y)\hspace{-1pt}=\hspace{-1pt}\frac{1}{{K + {{\sigma _g^2} \mathord{\left/
 {\vphantom {{\sigma _g^2} {\sigma _p^2}}} \right.
 \kern-\nulldelimiterspace} {\sigma _p^2}}}}\sum\nolimits_{k = 0}^{K - 1} {z_k(x,y)},\hspace{2pt} {\forall(x,y)}\in\Omega,
\end{equation}
which averages the whole cluster $\{z_k\}_{k=0}^{K-1}$. From the expression of ${\widetilde{\bf{g}}}$ \eqref{eq:sta_fin}, ${f}_3^\star(x,y)$ corresponds to the term ${\widetilde{\bf{g}}}{\widetilde{\bf{g}}} ^{\rm{T}}$ in \eqref{eq:G_closed}. The discretized version of \eqref{eq:fea_inten} is
\begin{equation}\label{eq:fea_inten_2}
{\bf{f}}_3^\star\hspace{-1pt}=\hspace{-1pt}\frac{1}{{K + {{\sigma _g^2} \mathord{\left/
 {\vphantom {{\sigma _g^2} {\sigma _p^2}}} \right.
 \kern-\nulldelimiterspace} {\sigma _p^2}}}}\sum\nolimits_{k = 0}^{K - 1} {{{\bf{z}}_k}}.
\end{equation}
With the defined ${{\bf{f}}_1^\star}$, ${{\bf{f}}_2^\star}$ and ${{\bf{f}}_3^\star}$, we can obtain the neighborhood graph $\mathcal{G}^\star$, and hence its graph Laplacian $\mathbf{L}^\star$ and graph Laplacian regularizer $S_{\mathcal{G}^\star}$, as discussed in Section\;\ref{sec:interp_xlx}.

Note that from \eqref{eq:fea_spat} and \eqref{eq:fea_inten_2}, ${\bf{f}}_1^\star$ and ${\bf{f}}_2^\star$ reflect spatial relationship while ${\bf{f}}_3^\star$ is related to pixel intensities. 
Such a setting is, at a glance, similar to that of bilateral filtering \cite{tomasi98}.
However, our work not only operates \emph{non-locally} but also \emph{optimally} balances the contributions from the spatial and intensity components, leading to superior denoising performance.

%% file: diffu_xlx.tex
Based on the convergence result in Section \ref{ssec:conv} and the optimal metric space derived in Section \ref{ssec:opt_met}, we now delineate the fundamental relationship between graph Laplacian regularization and anisotropic diffusion\cite{weickert98,perona90}. 
The interpretation in this section gives more insights into the behavior of graph Laplacian regularization, {\it e.g.}, its tendency to promote piecewise smooth results under certain conditions.

\subsection{Graph Laplacian Regularization as Tensor Diffusion}\label{ssec:to_diffu}
We first show that graph Laplacian regularization can be interpreted as an anisotropic tensor diffusion scheme. With the convergence of the graph Laplacian regularizer $S_\mathcal{G}$ to the functional $S_\Omega$ \eqref{eq:confunc}, the continuous counterpart of the denoising problem \eqref{eq:prob_discrete} is given by
\begin{equation}\label{eq:prob_contin}
{u^ \star } = \mathop {\arg \min }\limits_u \left\| {u - {z_0}} \right\|_\Omega ^2 +\tau\hspace{-1pt}\bm\cdot\hspace{-3pt}\int_\Omega {{{\nabla u}^{\rm{T}}}{{\bf{D}}}\nabla u} \; d{\bf{s}},
\end{equation}
where we denote ${\bf{D}}={{\bf{G}}^{ - 1}}{{\left( {\sqrt {\det {\bf{G}}} } \right)}^{2\gamma  - 1}}$ for simplicity.\footnote{The value of $\tau$ in \eqref{eq:prob_contin} is different from that in \eqref{eq:prob_discrete} because from \eqref{eq:conv}, ${\cal S}_{\cal G}({\bf u})$ converges to ${\cal S}_{\Omega}(u)$ up to a scaling factor.}
Like ${\bf G}$, the newly defined ${\bf D}:\Omega\mapsto\mathbb{R}^{2\times 2}$ is also a matrix-valued function of ${\bf s}\in\Omega$. 
To solve \eqref{eq:prob_contin}, we differentiate its objective with respect to $u$, and then equate it to zero:
\begin{equation}\label{eq:prob_sol}
u^\star=z_0+\tau\,{\rm{div}}\left( {{\bf D}\nabla u^\star} \right).
\end{equation}
Similar to the derivation in \cite{strong96}, the denoised patch $u^\star$ can be obtained by running the following diffusion scheme \emph{forward} in time on noisy patch $z_0$ with step size $\tau$:
\begin{align}
{\partial _t}u &= {\rm{div}}\left( {{\bf D}\nabla u} \right),\label{eq:prob_diffu}\\
u({\bf{s}},t = 0) &= {z_0}({\bf{s}}).\label{eq:diffu_init}
\end{align}
In other words, \textit{marching} the patch $z_0$ forward in time using the diffusion equation \eqref{eq:prob_diffu} with step size $\tau$ results in $u^\star$.
In \eqref{eq:prob_diffu}, $u$ is a 3D function of space and time, {\it i.e.}, $u({\bf s},t):\Omega\times[0,\tau]\mapsto\mathbb{R}$. 
Hence $\nabla u=\left[ {{\partial _x}u{\textrm{  }}{\partial _y}u} \right]^{\rm T}:\Omega\times[0,\tau]\mapsto\mathbb{R}^2$ is a vector-valued function of space and time.
In \eqref{eq:prob_diffu}, the quantity multiplying $\nabla u$\----called the \emph{diffusivity} \cite{weickert98}\----is the 2D tensor ${\bf D}$, which determines how fast the image $u$ is diffused.
As a result, \eqref{eq:prob_diffu} belongs to a class of anisotropic diffusion schemes called \emph{tensor diffusion}~\cite{weickert96}.
We now see that graph Laplacian regularization is the discrete counterpart of time-marching the noisy image using an anisotropic tensor diffusion scheme with tensor ${\bf{D}}$. 

We note that several existing diffusion schemes, {\it e.g.,}\cite{weickert94,perona90,strong96}, are special cases of \eqref{eq:prob_diffu}.
We herein focus on analyzing \eqref{eq:prob_diffu}, with the diffusion tensor $\bf{D}^\star$ derived from the optimal metric space ${\bf{G}}^\star$ \eqref{eq:G_closed}, {\it i.e.}, ${\bf{D}}^\star={{{\bf{G}}^\star}^{ - 1}}{{\left( {\sqrt {\det {\bf{G}}^\star} } \right)}^{2\gamma  - 1}}$.
Hence ${\bf D}^\star$ is dependent on the noisy patch $z_0$. In this case, \eqref{eq:prob_diffu} is called a \emph{nonlinear} diffusion because its diffusivity is a function of the current observation \cite{burgers13}.
With \eqref{eq:G_eig}, \eqref{eq:G_eig_2}, and the optimal metric space \eqref{eq:G_closed}, the tensor ${\bf D}^\star$ can be eigen-decomposed as
\begin{equation}\label{eq:diffu_tensor_0}
{\bf{D}}^\star = {\beta_{\mathcal{G}}^{2\gamma  - 1}}
\begin{bmatrix} 
{{\bf{v}}_1} & {{\bf{v}}_2}
\end{bmatrix}
\begin{bmatrix} 
{\lambda _1} & 0\\
0 & {\lambda _2}
\end{bmatrix}
\begin{bmatrix} 
{{\bf{v}}_1} & {{\bf{v}}_2}
\end{bmatrix}^{\rm{T}}.
\end{equation}
Recall that $\beta_{\mathcal{G}}={\widetilde{\sigma}}^2+\beta$. 
${\bf{v}}_1$ and ${\bf{v}}_2$ are unit vectors corresponding to the two columns of matrix $\bf{U}$ in \eqref{eq:G_eig_2}. 
Their directions are related to that of the estimated gradient ${\widetilde{\bf g}}$, where ${\bf{v}}_1$ is \emph{parallel} to ${\widetilde{\bf g}}$ and ${\bf{v}}_2$ is \emph{perpendicular} to ${\widetilde{\bf g}}$.
In addition, one can derive that eigenvalues $\lambda_1$ and $\lambda_2$ are scalar functions of ${{{\left\| {\widetilde {\bf{g}}} \right\|}_2}}$:
\begin{equation}\label{eq:diffu_tensor_1}
{\lambda _1}({\left\| {\widetilde {\bf{g}}} \right\|_2})\hspace{-2pt}=\hspace{-2pt}{\left(\hspace{-3pt}{1 + \frac{{\left\| {\widetilde {\bf{g}}} \right\|_2^2}}{\beta_{\mathcal{G}}^2}} \right)^{\gamma  - 1.5}}\hspace{-3pt}, 
{\lambda _2}({\left\| {\widetilde {\bf{g}}} \right\|_2})\hspace{-2pt}=\hspace{-2pt}{\left(\hspace{-3pt}{1 + \frac{{\left\| {\widetilde {\bf{g}}} \right\|_2^2}}{\beta_{\mathcal{G}}^2}} \right)^{\gamma  - 0.5}}.
\end{equation}
From \eqref{eq:diffu_tensor_1}, $\lambda_1/\lambda_2<1$ holds for any ${\left\| {\widetilde {\bf{g}}} \right\|_2}>0$, and $\lambda_1=\lambda_2=1$ for ${\left\| {\widetilde {\bf{g}}} \right\|_2}=0$. 
According to \cite{weickert98}, these imply that the diffusion equation \eqref{eq:prob_diffu} with tensor ${\bf D}^\star$ is \emph{edge-preserved}.\footnote{We refer interested readers to \cite{weickert98} for a more detailed treatment of the properties of tensor diffusion.}

Given the decomposition of ${\bf D}^\star$ in \eqref{eq:diffu_tensor_0} and \eqref{eq:diffu_tensor_1}, we now simplify the tensor diffusion equation \eqref{eq:prob_diffu} to one with \emph{scalar} diffusivity, also known as \emph{Perona-Malik diffusion} \cite{perona90}. 
By doing so, we can introduce the notions of forward and backward diffusion, so as to explain the behavior of graph Laplacian regularization under different settings.

\subsection{Graph Laplacian Regularization as Perona-Malik Diffusion}
\label{ssec:pws}
%

Suppose the noise variance $\sigma_g^2$ is small. 
Then, first, $\widetilde{{\bf{g}}}\approx \bf{g}$ from \eqref{eq:noise} and \eqref{eq:sta_fin}, {\it i.e.}, the gradient estimate is close to the ground-truth. 
Second, for effective denoising, $\nabla u$ should approach the ground-truth ${\bf g}$ when diffusing using \eqref{eq:prob_diffu}, {\it i.e.,} ${\bf g}\approx\nabla u$.
Consequently, $\widetilde{\bf{g}}\approx\nabla u$ when $\sigma_g^2$ is small.
In fact, we perform denoising iteratively with decreasing noise (Section\,\ref{sec:alg}), so at least for the last few iterations, the noise variance $\sigma_g^2$ is small and $\widetilde{\bf{g}}$ should be close to $\nabla u$.

By letting $\widetilde{{\bf{g}}}=\nabla u$ in \eqref{eq:diffu_tensor_0} and \eqref{eq:diffu_tensor_1}, we can simplify the diffusion equation \eqref{eq:prob_diffu} to
\begin{equation}\label{eq:diffu_pm}
{\partial _t}u = {\beta_{\mathcal{G}}^2}\hspace{3pt}{\rm{div}}\left( {{\lambda _1}({{\left\| {\nabla u} \right\|}_2})\nabla u} \right),
\end{equation}
which is the Perona\--Malik diffusion \cite{perona90} with ${\lambda _1}({{\left\| {\nabla u} \right\|}_2})$ as the scalar diffusivity. 
Next we decompose \eqref{eq:diffu_pm} into two diffusion processes and present the notions of forward and backward diffusion for detailed analysis.

We first define a scalar function $J_1(\bm\cdot)$ of ${{\left\| {\nabla u} \right\|}_2}$:
\begin{equation}\label{eq:flux}
J_1({{\left\| {\nabla u} \right\|}_2})=\lambda_1({{\left\| {\nabla u} \right\|}_2}){{\left\| {\nabla u} \right\|}_2},
\end{equation}
which is the magnitude of the vector $\rm{div}(\bm\cdot)$ is operating on in \eqref{eq:diffu_pm}.
It is also called the \emph{flux function} in the literature \cite{perona90,weickert98}.
According to \cite{alvarez93}, \eqref{eq:diffu_pm} can be rewritten as
\begin{equation}\label{eq:diffu_gauge}
{\partial _t}u = \beta _{\cal G}^2\bm\cdot\Big( {{\lambda _1}\left( {{{\left\| {\nabla u} \right\|}_2}} \right){{\partial _{\zeta \zeta }}u} + {J'_{1}}\left( {{{\left\| {\nabla u} \right\|}_2}} \right){\partial _{\eta \eta }}u} \Big),
\end{equation}
where $J_1'({{\left\| {\nabla u} \right\|}_2})$ is the derivative of $J_1({{\left\| {\nabla u} \right\|}_2})$ with respect to ${{\left\| {\nabla u} \right\|}_2}$. 
$\zeta$ and $\eta$ are called \emph{gauge coordinates} and denote the directions \emph{perpendicular} and \emph{parallel} to gradient $\nabla u$, respectively. 
${\partial _{\zeta \zeta }}u$ is the second order derivative of $u$ in the direction of $\zeta$, which indicates a diffusion process perpendicular to $\nabla u$ (or along edges). 
The scalar function multiplying ${\partial _{\zeta \zeta }}u$, {\it i.e.,} $\lambda_1({{\left\| {\nabla u} \right\|}_2})$, is the diffusivity determining how fast $u$ is diffused along edges. 
Similarly, ${\partial _{\eta \eta }}u$ represents a diffusion process across edges, and $J_1'({{\left\| {\nabla u} \right\|}_2})$ determines how fast $u$ is diffused across edges.

\begin{figure}[!t]
\centering
    \subfigure[]{\includegraphics[height=75pt]{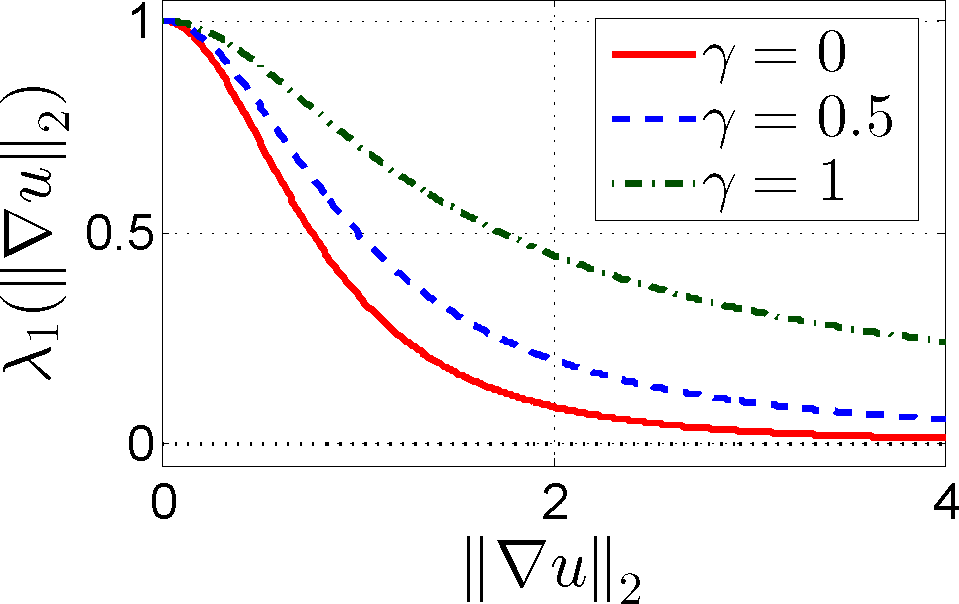}\label{fig:flux_a}}\hspace{8pt}
    \subfigure[]{\includegraphics[height=75pt]{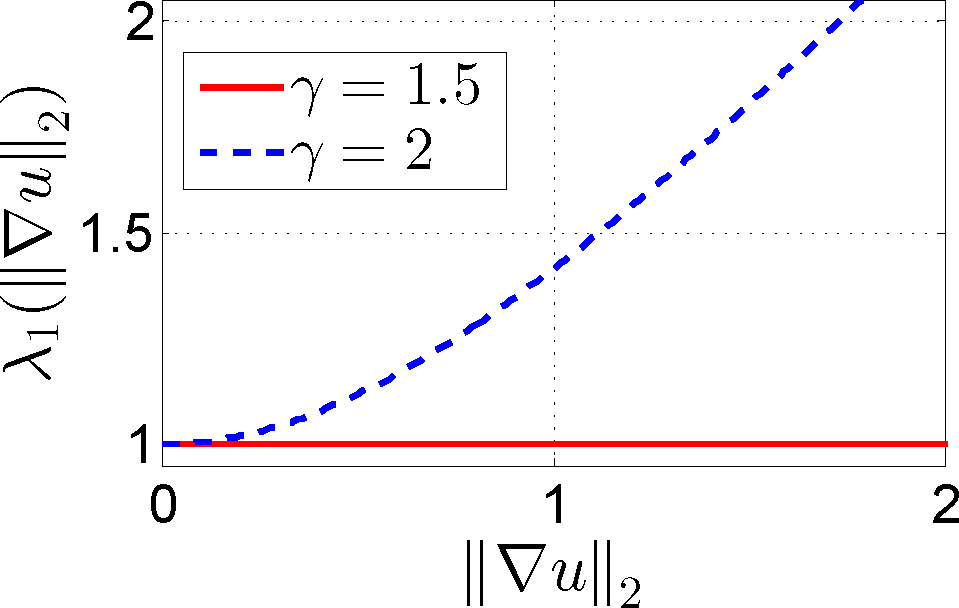}\label{fig:flux_b}}\vspace{-3pt}\\
    \subfigure[]{\includegraphics[height=75pt]{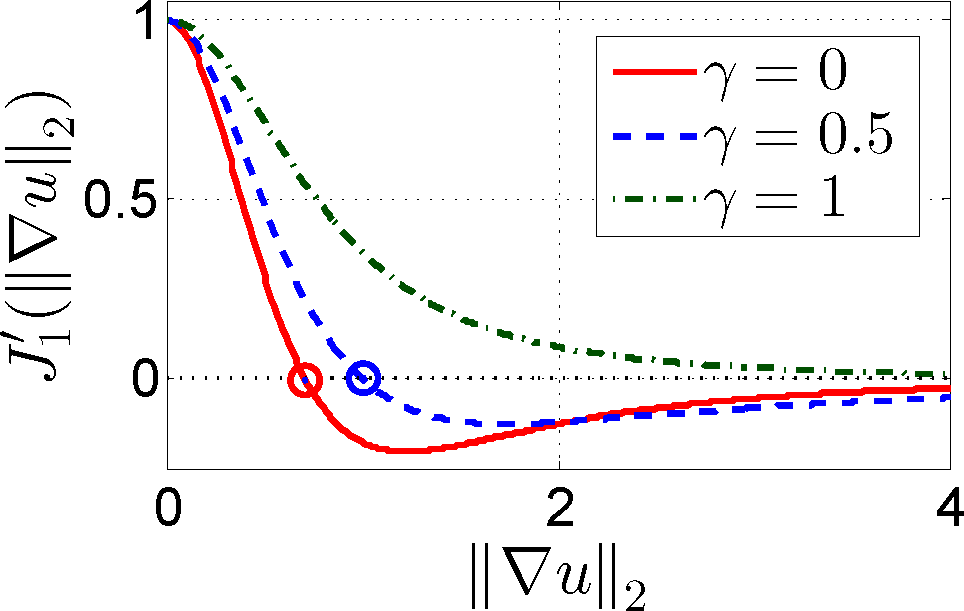}\label{fig:flux_c}}\hspace{8pt}
    \subfigure[]{\includegraphics[height=75pt]{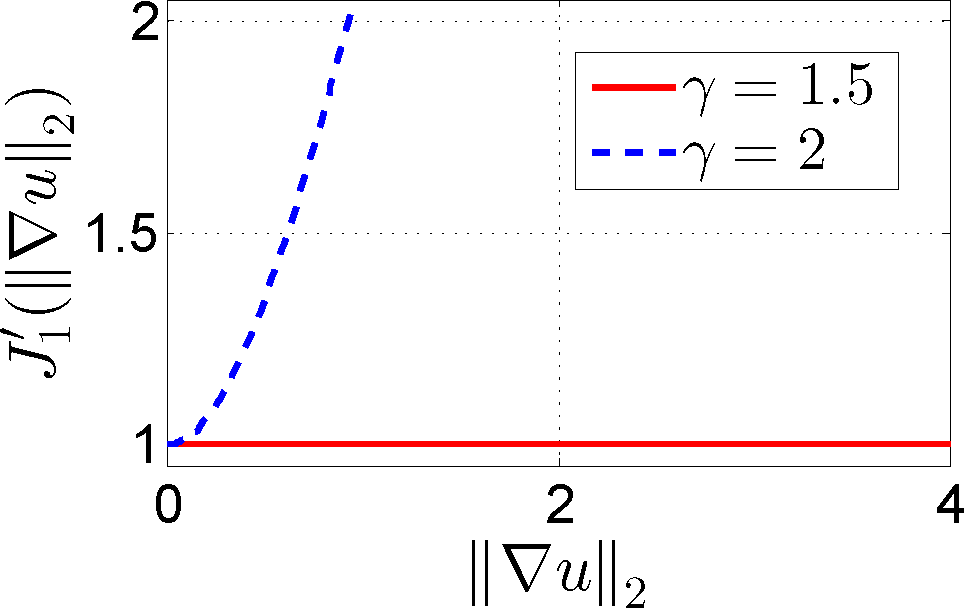}\label{fig:flux_d}}\vspace{-3pt}
\caption{Diffusivities of \eqref{eq:diffu_gauge} along the edges (in (a) and (b)) and across the edges (in (c) and (d)) with different $\gamma$'s. The circles in (c) mark the contrast parameter $T$. For illustration, we set $\beta_{\cal G}^2=1$ when plotting these diagrams.}
\label{fig:flux}
\end{figure}

We see that \eqref{eq:diffu_gauge} decouples \eqref{eq:diffu_pm} into two \emph{independent} diffusion processes: one \emph{along} edges with diffusivity $\lambda_1\left( {{{\left\| {\nabla u} \right\|}_2}} \right)$ and the other one \emph{across} edges with diffusivity $J'_1\left( {{{\left\| {\nabla u} \right\|}_2}} \right)$. 
From \eqref{eq:diffu_tensor_1}, $\lambda_1\left( {{{\left\| {\nabla u} \right\|}_2}} \right)>0$ always holds, no matter what value the normalization parameter $\gamma$ takes.
For example, Fig.\,\ref{fig:flux_a} and Fig.\,\ref{fig:flux_b} plot several curves of $\lambda_1\left( {{{\left\| {\nabla u} \right\|}_2}} \right)$ as a function of ${\left\| {\nabla u} \right\|}_2$ for different $\gamma\in[0,2]$, and we see that the value of $\lambda_1\left( {{{\left\| {\nabla u} \right\|}_2}} \right)$ is always positive.
According to \cite{alvarez93}, a positive $\lambda_1\left( {{{\left\| {\nabla u} \right\|}_2}} \right)$ means that \eqref{eq:diffu_gauge} always has a \emph{forward diffusion}, {\it{i.e.}}, blurring/smoothing process, along edges. 
However, the diffusivity across edges is $J'_1\left( {{{\left\| {\nabla u} \right\|}_2}} \right)$.
From \eqref{eq:diffu_tensor_1} and \eqref{eq:flux}, we can derive
\begin{equation}
{J'_1}({{{\left\| {\nabla u} \right\|}_2}})\hspace{-3pt}=\hspace{-3pt}{\left( \hspace{-3pt}{1\hspace{-2pt}+\hspace{-2pt}\frac{{{{{{\left\| {\nabla u} \right\|}_2^2}}}}}{{\beta _{\cal G}^2}}} \right)^{\hspace{-4pt}\gamma - 2.5}}\hspace{-2pt}\bm\cdot\hspace{-2pt}\left(\hspace{-1pt}{1\hspace{-2pt}+\hspace{-2pt}\frac{{2{{{{\left\| {\nabla u} \right\|}_2^2}}}}}{{\beta _{\cal G}^2}}\left( {\gamma \hspace{-2pt}-\hspace{-2pt}1} \right)}\hspace{-2pt}\right).
\end{equation}
It behaves differently according to different choices of the normalization parameter $\gamma$.

\subsubsection{Forward-backward diffusion when $\gamma<1$} 
We first define a constant
\begin{equation}\label{eq:thres}
T = {{{\beta _G}} \mathord{\left/
 {\vphantom {{{\beta _G}} {\sqrt {2(1 - \gamma )} }}} \right.
 \kern-\nulldelimiterspace} {\sqrt {2(1 - \gamma )} }}.
\end{equation}
Given $\gamma<1$, one can show that $J'_1({{{\left\| {\nabla u} \right\|}_2}})>0$ for ${{{\left\| {\nabla u} \right\|}_2}}<T$, and $J'_1({{{\left\| {\nabla u} \right\|}_2}})\le0$ for ${{{\left\| {\nabla u} \right\|}_2}}\ge T$. 
For example, we can see the curves of $J'_1({{{\left\| {\nabla u} \right\|}_2}})$ as a function of ${\left\| {\nabla u} \right\|}_2$ for $\gamma\in\{0,0.5\}$ in Fig.\;\ref{fig:flux_c}, where the circles mark the positions of $T$ with different $\gamma$'s. 
Thus, we can conclude:
\begin{enumerate}[(i)]
\item{If gradient ${\left\| {\nabla u} \right\|_2}<T$, then $ J'_1({{{\left\| {\nabla u} \right\|}_2}})>0$ and there is a \emph{forward diffusion} (smoothing) across edges;}
\item{If gradient ${{{\left\| {\nabla u} \right\|}_2}}=T$, then the diffusivity is $J'_1({{{\left\| {\nabla u} \right\|}_2}})\hspace{-1pt}=\hspace{-1pt}0$, so there is \emph{no diffusion} across the edges;}
\item{If gradient ${{{\left\| {\nabla u} \right\|}_2}}>T$, then $J'_1({{{\left\| {\nabla u} \right\|}_2}})<0$. This negative diffusivity means there is a \emph{backward diffusion} across edges. From \cite{weickert98}, it inverts the heat equation locally, leading to enhanced/sharpened edges.}
\end{enumerate}
Consequently, edges with ${{{\left\| {\nabla u} \right\|}_2}}\ge T$ are either maintained, or enhanced by backward diffusion; while smooth regions with ${{{\left\| {\nabla u} \right\|}_2}}<T$ are blurred by forward diffusion.
This phenomenon is called \emph{forward-backward} diffusion \cite{weickert98}, and it promotes \emph{piecewise smooth} results, as noted in the works \cite{perona90,weickert98}, and \cite{alvarez93}. 
As will be shown in the experimentation (Section \ref{sec:results}), a small $\gamma$ ({\it e.g.}, $\gamma=0$) is particularly useful for recovering piecewise smooth images, though it may create false edges.

The constant $T$ \eqref{eq:thres} is called the \emph{contrast parameter} \cite{weickert94,weickert98}, and it separates forward diffusion and backward diffusion.
From \eqref{eq:thres}, a smaller $\gamma$ would lead to a smaller $T$, {\it e.g.}, see the positions of $T$ in Fig.\,\ref{fig:flux_c} marked by the circles. 
Therefore a smaller $\gamma$ makes the backward diffusion occur more easily, leading to more edge enhancement of an image.

\subsubsection{Relation to TV regularization when $\gamma=1$}
%
From \cite{strong96} and \cite{chan99}, TV regularization that minimizes the functional $\int_\Omega {{{\left\| {\nabla u} \right\|}_2}}\,d{\bf{s}}$ is equivalent to time-marching an image using the following diffusion scheme:
\begin{equation}\label{eq:diffu_tv}
{\partial _t}u = {\rm{div}}\left( {{{\left( {\left\| {\nabla u} \right\|_2^2 + \beta _{\rm TV}^2} \right)}^{ - 0.5}}\nabla u} \right),
\end{equation}
where ${\beta_{\rm TV}}$ is a positive constant to ensure \eqref{eq:diffu_tv} is well-defined when $\nabla u={\bf 0}$. 
By letting $\gamma=1$, \eqref{eq:diffu_pm} can be rewritten as 
\begin{equation}\label{eq:gamma_one}
{\partial _t}u = \beta _{\cal G}^3\bm\cdot{\rm{div}}\left( {{{\left( {\left\| {\nabla u} \right\|_2^2 + \beta _{\cal G}^2} \right)}^{ - 0.5}}\nabla u} \right).
\end{equation}
One can clearly see the similarity between \eqref{eq:gamma_one} and \eqref{eq:diffu_tv}.
Therefore graph Laplacian regularization can be viewed as a \emph{discretization} of TV regularization when $\gamma=1$.

In this case, the diffusivity across edges, {\it i.e.}, ${{J_1'}}\left( {{{\left\| {\nabla u} \right\|}_2}} \right)$, is always positive, and ${{J_1'}}\left( {{{\left\| {\nabla u} \right\|}_2}} \right)\rightarrow0$ as ${{{\left\| {\nabla u} \right\|}_2}}\rightarrow+\infty$; see the curve for $\gamma=1$ in Fig.\,\ref{fig:flux_c}.
It means TV regularization (or the special case where $\gamma=1$ for graph Laplacian regularization) can neither enhance edges nor eliminate sharp transitions. 
As mentioned in \cite{strong96}, it is a \emph{canonical} case of geometry-driven diffusion, which limits its usage for images with different characteristics. Moreover, TV is a local method, as mentioned in Section\,\ref{ssec:gsp_review}; while our proposal incorporates non-local information for effective denoising.

\subsubsection{Forward diffusion when $\gamma>1$}
In this case, we have $J'_1({{{\left\| {\nabla u} \right\|}_2}})>0$ ({\it e.g.}, the curves in Fig.\,\ref{fig:flux_d}) and there is always a forward diffusion to blur the edges, which is not conducive to recovering image structures. 
However, this case never creates false edges as there is no edge enhancement.

From the above analysis, we see that $\gamma$ determines the types of diffusion that can occur, which leads to different denoising effects. 
Our work gives users the freedom to choose the appropriate $\gamma$, according to different types of images to be restored. 
The denoised results under different $\gamma$'s will be presented and discussed in Section \ref{sec:results}.

%% file: alg.tex
To demonstrate the practicality of our previous analysis, we develop an iterative patch-based image denoising algorithm.
Given a noisy image ${\cal I}$ (corrupted by i.i.d. AWGN in the pixel domain) and its noise variance $\sigma_{\cal I}^2$, our algorithm denoises ${\cal I}$ with graph Laplacian regularization in an iterative manner.
For convenience, we also denote ${\cal I}^{(0)}={\cal I}$ and $\sigma_{\cal I}^{(0)}=\sigma_{\cal I}$.
Our method is called \emph{optimal graph Laplacian regularization} (OGLR) for denoising. 
We summarize our method in Algorithm\;\ref{alg:denoi}, and its key steps are elaborated as follows.
\begin{algorithm}[t]
\caption{Image denoising with OGLR}\label{alg:denoi}
\begin{small}
\begin{algorithmic}[1]
\STATE {\bf{Input:}} Noisy image ${\cal I}$, noise variance $\sigma_{\cal I}^2$
\FOR {$k=0$ to $iter-1$}
\FOR {each noisy patch ${\bf z}_0$}
\STATE Clustering of similar patches on ${\cal I}^{(k)}$
\STATE Computation of graph Laplacian from similar patches
\STATE Denoising of ${\bf z}_0$ with constrained optimization
\ENDFOR
\STATE Aggregation of the denoised image ${\cal I}^{(k+1)}$
\IF {$(\sigma_{\cal I}^{(k)})^2\ge\sigma_{\rm th}^2$ and $k\neq iter-1$}
\STATE Estimation of noise variance $(\sigma _{\cal I}^{(k + 1)})^2$
\ELSE
\STATE {\bf return} ${\cal I}^{(k+1)}$
\ENDIF
\ENDFOR
\STATE {\bf{Output:}} Denoised image ${\cal I}^{(k+1)}$
\end{algorithmic}
\end{small}
\end{algorithm}
\subsection{Clustering of Similar Patches}\label{ssec:alg_cluster}
We denoise one-by-one size $\sqrt{M}\times\sqrt{M}$ pixel patches, spaced $N_S$ pixels apart in the noisy image. 
The value $N_S$, which determines the amount of patch overlaps, is set differently according to different noise variances. 
To denoise each patch ${\bf z}_0$, we first search for its $K-1$ most similar patches, where the patch distances are measured after \emph{coarse pre-filtering}, as similarly done in BM3D\,\cite{dabov07}. Specifically: 
\begin{enumerate}[(i)]
\item We first transform ${\bf z}_0$ into the 2D-DCT domain, and then apply hard-thresholding to the spectral coefficients. By transforming it back to the spatial domain, we obtain the filtered patch, denoted as $\Upsilon({\bf z}_0)$.
\item From the noisy image, we search for the $K-1$ patches $\{{\bf z}_k\}_{k=1}^{K-1}$ that are most similar to ${\bf z}_0$, where we use the Euclidean distance as metric and measure the distances with the filtered patches; {\it e.g.}, the distance between ${\bf z}_0$ and a candidate patch ${\bf z}_{\rm can}$ is ${\left\| {\Upsilon ({{\bf{z}}_0}) - \Upsilon ({{{\bf{z}}}_{\rm can}})} \right\|_2}$.\footnote{Because DCT is an orthonormal transform, for simplicity, our implementation does not transform the patches back to the spatial domain and computes the patch distances based on the spectral coefficients directly.}
\end{enumerate}
Having found cluster of similar patches $\{{\bf z}_k\}_{k=0}^{K-1}$, we denoise ${\bf z}_0$ in the following steps.
%
\subsection{Graph Laplacian from Similar Patches}
Given a noisy target patch ${\bf z}_0$ and its similar cluster $\{{\bf{z}}_k\}_{k=0}^{K-1}$, we next compute the optimal graph Laplacian ${\bf L}^\star$ for recovering ${\bf z}_0$.
We first need to estimate $\sigma_g^2$---the variance of the noisy gradients in the continuous domain---from cluster $\{{\bf{z}}_k\}_{k=0}^{K-1}$ that includes the target patch.
To do so, we compute the discrete gradients of patches $\{{\bf{z}}_k\}_{k=0}^{K-1}$ with two filters, ${\bf{F}}_x=[1\ -\hspace{-1pt}1]$ and ${\bf{F}}_y=[1\ -\hspace{-1pt}1]^{\rm{T}}$, leading to 2D gradients ${\bf{g}}_k^{(i)}=[g_{x,k}^{(i)}\hspace{3pt}g_{y,k}^{(i)}]^{\rm T}$ for $0\le\hspace{-1pt}k\hspace{-1pt}\le\hspace{-1pt}K\hspace{-1pt}-\hspace{-1pt}1$ and $1 \le i \le M$. 
Then for each pixel $i$, we compute the sample variance of $\{g_{x,k}^{(i)}\}_{k=0}^{K-1}$ and the sample variance of $\{g_{y,k}^{(i)}\}_{k=0}^{K-1}$, respectively. Since every patch has $M$ pixels, we obtain $2M$ variances.
We empirically set $\sigma_g^2$ to be the average of all these $2M$ variances times a constant $\nu$. 
With the estimated $\sigma_g^2$, we compute the discrete exemplar functions $\{{\bf f}_1^\star, {\bf f}_2^\star, {\bf f}_3^\star\}$ with \eqref{eq:fea_spat} and \eqref{eq:fea_inten_2}, leading to the edge weights and graph Laplacian ${\bf L}^\star$, as presented in Section\,\ref{ssec:graph}.
\subsection{Patch-Based Denoising with Constrained Optimization}\label{ssec:denoi_patch}
Having obtained the optimal graph Laplacian ${\bf L}^\star$, the target patch ${\bf{z}}_0$ is denoised in this step, via the following constrained formulation:
%
\begin{equation}\label{eq:prob_constrained}
{{\bf{u}}^ \star } = \mathop {\arg \min }\limits_{\bf{u}} {{\bf{u}}^{\rm{T}}}{\bf{L}}^\star{\bf{u}}\hspace{8pt}{\rm{s.t.}}\hspace{6pt}\left\| {{\bf{u}} - {{\bf{z}}_0}} \right\|_2^2 = {C_{\cal I}}({\sigma _{\cal I}^{(k)}})^2,
\end{equation}
where $(\sigma_{\cal I}^{(k)})^2$ is the noise variance of the noisy image ${\cal I}^{(k)}$ and $C_{\cal I}$ is a constant controlling the proportion of noise to be removed. 
Note that solving \eqref{eq:prob_constrained} is \emph{equivalent} to solving \eqref{eq:prob_discrete} with $\tau=2/\delta$, where $\delta$ is the Lagrange multiplier found when solving \eqref{eq:prob_constrained}. 
Hence our analysis developed for \eqref{eq:prob_discrete} is also applicable for \eqref{eq:prob_constrained}. This methodology of choosing the regularization strength is called the \emph{discrepancy principle} in the literature \cite{scherzer93,strong96}.
The problem \eqref{eq:prob_constrained} is a quadratically constrained quadratic programming (QCQP) problem; it is convex and can be efficiently solved, {\it e.g.,} based on a Newton method, as discussed in \cite{hansen94}.

If the noisy image is to be recovered in only one iteration, then $C_{\cal I}$ is set to be close to $1$. 
However, similar to existing methods, {\it e.g.,}\cite{elad06,hu13} and \cite{dabov07}, we perform denoising iteratively, as suggested by \cite{milanfar13}, so as to achieve a better performance. 
Consequently, we let $C_{\cal I}$ be less than $1$, {\it e.g.,} 0.7, and denoise the target patch ${\bf z}_0$ with \eqref{eq:prob_constrained}. 
Hence, part of the noise remains in future iterations. 
The value of $C_{\cal I}$ is set close to $1$ only if the current noise variance $(\sigma_{\cal I}^{(k)})^2$ is smaller than a threshold $\sigma_{\rm th}^2$ or the maximum number of iterations is reached. 
By doing so, we can remove all of the remaining noise in the last iteration. 

\subsection{Denoised Image Aggregation}
Having obtained all the denoised overlapping patches, we aggregate all of them to form the denoised image ${\cal I}^{(k+1)}$.
Specifically, each pixel of ${\cal I}^{(k+1)}$ is estimated as the weighted average of the values from different overlapping patches. 
If a patch ${\bf z}_0$ has similar patches $\{{\bf z}_k\}_{k=1}^{K-1}$ with strong similarity to ${\bf z}_0$, then we expect that ${\bf z}_0$ can be restored to a high quality. 
Consequently, we empirically set the weight of a denoised ${\bf z}_0$ to be inversely proportional to $\sum\nolimits_{k = 1}^{K - 1} {\left\| {\Upsilon({{\bf{z}}_k}) - \Upsilon({{\bf{z}}_0})} \right\|_2^2}$. Recall that $\Upsilon(\bm\cdot)$ is the pre-filtering operator of patch clustering described in Section\,\ref{ssec:alg_cluster}.
\subsection{Noise Level Estimation}
We estimate the noise variance of image ${\cal I}^{(k+1)}$, {\it i.e.}, $(\sigma_{\cal I}^{(k+1)})^2$, before proceeding to the next iteration.
Denote the total number of pixels in the image as $N_{\cal I}$. 
We also use ${\cal I}^{(k)}$ and ${\cal I}^{(k+1)}$ to represent their respective vectorized images, and hence ${\cal I}^{(k)},{\cal I}^{(k+1)}\in\mathbb{R}^{N_{\cal I}}$. 
For simplicity, we herein adopt a collinear assumption---assuming the noiseless (original) image, ${\cal I}^{(k)}$, and ${\cal I}^{(k+1)}$ are three points on the same line in the high dimensional space $\mathbb{R}^{N_{\cal I}}$. Then we can derive
\begin{equation}\label{eq:noi_est}
\sqrt {N_{\cal I}} {\sigma _{\cal I}^{(k+1)}} = \sqrt {N_{\cal I}} {\sigma _{\cal I}^{(k)}} - {\left\| {{\cal I}^{(k)} - {\cal I}^{(k+1)}} \right\|_2}.
\end{equation}
With \eqref{eq:noi_est}, the new noise variance $(\sigma _{\cal I}^{(k+1)})^2$ can be obtained.
From \eqref{eq:prob_constrained}, the validity of the above collinear assumption mainly relies on two factors. 
First, we need an  effective graph Laplacian to promote the recovered patch towards the original one.
Second, we need a modest $C_{\cal I}$ to avoid over-smoothing---a big $C_{\cal I}$ always drives the denoised patch towards the DC. Since we not only construct the optimal graph Laplacian ${\bf L}^{\star}$ but also adopt a moderate $C_{\cal I}$ for recovery ($0.7$ in our case), it is reasonable to assume that the noiseless image, ${\cal I}^{(k)}$, and ${\cal I}^{(k+1)}$ are three collinear points. 
To be shown in Section\,\ref{sec:results}, our method provides satisfactory denoising performance, which also validates this collinear assumption empirically.

%% file: results.tex
\begin{table}[t]
  \centering\footnotesize
  \setlength{\tabcolsep}{3pt}
  \caption{Filtering of Two Noiseless Test Images Using Graph Laplacian Regularization with Different Normalizations.}
    \begin{tabular}{|c||c|c|c|}
    \hline
    \multirow{2}[4]{*}{\textbf{Image}} & \multicolumn{3}{c|}{\textbf{\bf $\phantom{\hat{I}}$Normalization Parameter $\gamma\phantom{\hat{I}}$}} \\
\cline{2-4}          & {{\vspace{-2.7pt}$\phantom{\hat{I}}\mathop{\textrm{0}}\limits_{\phantom{.}}\phantom{\hat{I}}$}} & 1 & 2 \\
    \hhline{====}
    \begin{minipage}[c]{55pt}\center
    \vspace{2pt}\includegraphics[height=55pt]{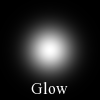}\vspace{2pt}
    \end{minipage} &
    \begin{minipage}[c]{55pt}\center
    \vspace{2pt}\includegraphics[height=55pt]{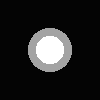}\vspace{2pt}
    \end{minipage} &
    \begin{minipage}[c]{55pt}\center
    \vspace{2pt}\includegraphics[height=55pt]{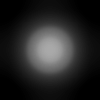}\vspace{2pt}
    \end{minipage} &
    \begin{minipage}[c]{55pt}\center
    \vspace{2pt}\includegraphics[height=55pt]{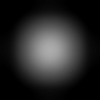}\vspace{2pt}
    \end{minipage}\\
    \hline
    \begin{minipage}[c]{55pt}\center
    \vspace{2pt}\includegraphics[height=55pt]{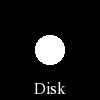}\vspace{2pt}
    \end{minipage} &
    \begin{minipage}[c]{55pt}\center
    \vspace{2pt}\includegraphics[height=55pt]{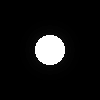}\vspace{2pt}
    \end{minipage} &
    \begin{minipage}[c]{55pt}\center
    \vspace{2pt}\includegraphics[height=55pt]{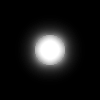}\vspace{2pt}
    \end{minipage} &
    \begin{minipage}[c]{55pt}\center
    \vspace{2pt}\includegraphics[height=55pt]{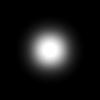}\vspace{2pt}
    \end{minipage} \\
    \hline
    \end{tabular}%
  \label{tab:exp_gamma}%
\end{table}%
%

We conducted extensive experiments to demonstrate the merits of our proposed denoising algorithm. 
Specifically, we investigate the impact of choosing different normalization parameter $\gamma$'s on the results. Then we evaluate our OGLR algorithm on denoising of natural images and piecewise smooth images, respectively.

\begin{figure*}[!t]
\centering
    \stackunder[5pt]{\includegraphics[width=80pt]{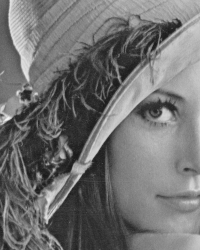}\hspace{2pt}\includegraphics[width=80pt]{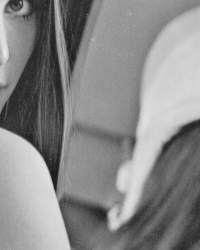}}{Original}\hspace{8pt}
    \stackunder[5pt]{\includegraphics[width=80pt]{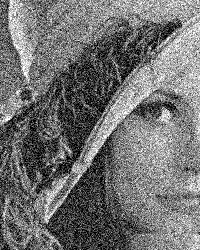}\hspace{2pt}\includegraphics[width=80pt]{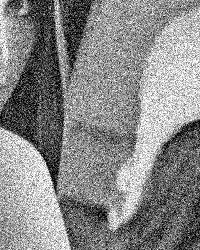}}{Noisy, 16.09\,dB}\hspace{8pt}
    \stackunder[5pt]{\includegraphics[width=80pt]{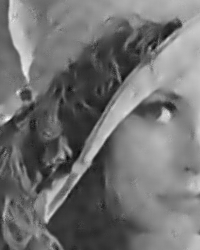}\hspace{2pt}\includegraphics[width=80pt]{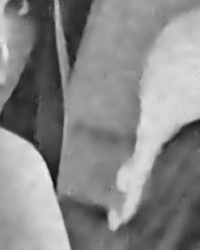}}{K-SVD, 29.02\,dB}\vspace{6pt}
    \stackunder[5pt]{\includegraphics[width=80pt]{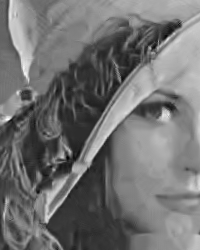}\hspace{2pt}\includegraphics[width=80pt]{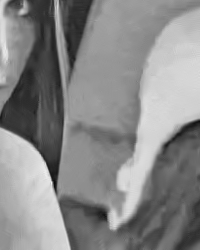}}{BM3D, 29.86\,dB}\hspace{8pt}
    \stackunder[5pt]{\includegraphics[width=80pt]{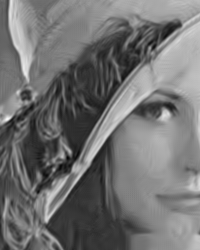}\hspace{2pt}\includegraphics[width=80pt]{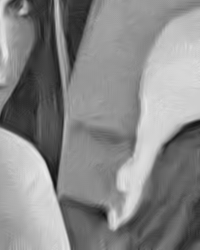}}{PLOW, 29.84\,dB}\hspace{8pt}
    \stackunder[5pt]{\includegraphics[width=80pt]{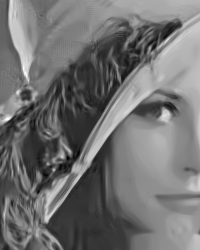}\hspace{2pt}\includegraphics[width=80pt]{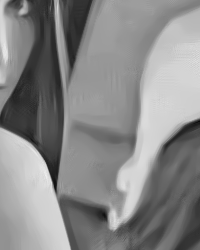}}{OGLR, 30.04\,dB}
\caption{Denoising of the natural image {\it Lena}, where the original image is corrupted by AWGN with $\sigma_{\cal I}=40$. Two cropped fragments of each image are presented for comparison.}
\label{fig:result_lena}
\end{figure*}
%

%
\begin{table*}[htbp]
  \centering\footnotesize
  \caption{Natural Image Denoising with OGLR: Performance Comparisons in
Average PSNR (Top, in dB) and SSIM Index (Bottom). In Each Cell, Results of Four Denoising Methods are Presented. Top Left: K-SVD \cite{elad06}. Top Right: BM3D \cite{dabov07}. Bottom Left: PLOW \cite{chatterjee12}. Bottom Right: OGLR (Proposed). The Best Results Among the Four Methods are Highlighted in Each Cell.}
\begin{tabular}{|c||c:c|c:c|c:c|c:c|c:c|c:c|c:c|}
    \hline
    \multirow{2}[4]{*}{\textbf{Image}} & \multicolumn{14}{l|}{\hspace{-2pt}${{\left[\begin{array}{c;{3pt/2pt}c}{\textrm{K-SVD}}&{\textrm{BM3D}}\\ \hdashline[3pt/2pt]{\mathop {\textrm{PLOW}}\limits^{}   }&{\textrm{OGLR}}\end{array}\right]}^{\phantom{\hat{a}}}_{\phantom{a}}}$\hspace{83pt}\textbf{\bf $\phantom{\hat{I}}$Standard Deviation $\sigma_{\cal I}\phantom{\hat{I}}$}} \\
\cline{2-15}          & \multicolumn{2}{c|}{\vspace{-2.7pt}$\phantom{\hat{I}}\mathop{\textrm{10}}\limits_{\phantom{.}}\phantom{\hat{I}}$} & \multicolumn{2}{c|}{20} & \multicolumn{2}{c|}{30} & \multicolumn{2}{c|}{40} & \multicolumn{2}{c|}{50} & \multicolumn{2}{c|}{60} & \multicolumn{2}{c|}{70} \\
    \hhline{===============}
    \multirow{4}[4]{*}{\textit{Lena}} & 35.55 & \textbf{35.89} & 32.40 & \textbf{33.02} & 30.42 & \textbf{31.23} & 28.96 & 29.82 & 27.80 & \textbf{29.00} & 26.87 & \textbf{28.20} & 26.11 & \textbf{27.50} \bigstrut[t]\\
          & 0.910 & \textbf{0.915} & 0.862 & \textbf{0.876} & 0.823 & \textbf{0.843} & 0.790 & 0.813 & 0.759 & \textbf{0.796} & 0.732 & \textbf{0.776} & 0.707 & \textbf{0.756} \\
\cdashline{2-15}          & 35.28 & 35.62 & 32.70 & 32.93 & 31.11 & 31.22 & 29.79 & \textbf{30.06} & 28.72 & 28.86 & 27.92 & 28.19 & 27.09 & 27.46 \bigstrut[t]\\
          & 0.906 & 0.912 & 0.871 & 0.874 & 0.842 & 0.842 & 0.809 & \textbf{0.821} & 0.776 & 0.785 & 0.752 & 0.768 & 0.719 & 0.742 \\
    \hline
    \multirow{4}[4]{*}{\textit{Barbara}} & 34.54 & \textbf{34.96} & 30.89 & \textbf{31.75} & 28.56 & \textbf{29.79} & 26.87 & 28.00 & 25.45 & 27.23 & 24.23 & 26.30 & 23.32 & 25.51 \bigstrut[t]\\
          & 0.936 & \textbf{0.942} & 0.881 & \textbf{0.905} & 0.821 & \textbf{0.867} & 0.767 & 0.822 & 0.714 & 0.794 & 0.662 & 0.759 & 0.617 & 0.727 \bigstrut[b]\\
\cdashline{2-15}          & 33.79 & 34.46 & 30.97 & 31.45 & 29.41 & 29.63 & 28.11 & \textbf{28.31} & 26.98 & \textbf{27.36} & 26.06 & \textbf{26.42} & 25.25 & \textbf{25.62} \bigstrut[t]\\
          & 0.928 & 0.937 & 0.892 & 0.902 & 0.860 & \textbf{0.867} & 0.823 & \textbf{0.838} & 0.783 & \textbf{0.801} & 0.746 & \textbf{0.768} & 0.710 & \textbf{0.734} \\
    \hline
    \multirow{4}[4]{*}{\textit{Peppers}} & 34.83 & \textbf{35.02} & 32.31 & \textbf{32.75} & 30.64 & \textbf{31.23} & 29.31 & 29.93 & 28.09 & \textbf{29.09} & 27.03 & \textbf{28.26} & 26.14 & \textbf{27.54} \bigstrut[t]\\
          & \textbf{0.879} & \textbf{0.879} & 0.839 & \textbf{0.845} & 0.811 & \textbf{0.820} & 0.786 & 0.795 & 0.762 & \textbf{0.782} & 0.738 & \textbf{0.763} & 0.715 & \textbf{0.746} \\
\cdashline{2-15}          & 34.40 & 34.91 & 32.40 & 32.67 & 31.01 & \textbf{31.23} & 29.80 & \textbf{30.10} & 28.76 & 28.83 & 27.86 & 28.20 & 27.17 & 27.42 \bigstrut[t]\\
          & 0.870 & \textbf{0.879} & 0.840 & 0.842 & 0.815 & 0.818 & 0.789 & \textbf{0.798} & 0.760 & 0.762 & 0.732 & 0.751 & 0.713 & 0.729 \\
    \hline
    \multirow{4}[4]{*}{\textit{Mandrill}} & 30.39 & \textbf{30.58} & 26.36 & \textbf{26.60} & 24.30 & \textbf{24.56} & 22.92 & 23.09 & 21.92 & 22.35 & 21.20 & 21.74 & 20.71 & 21.28 \bigstrut[t]\\
          & 0.895 & \textbf{0.897} & 0.778 & \textbf{0.792} & 0.675 & 0.702 & 0.582 & 0.617 & 0.503 & 0.549 & 0.443 & 0.498 & 0.401 & 0.459 \\
\cdashline{2-15}          & 29.58 & 29.84 & 26.10 & 26.35 & 24.33 & \textbf{24.56} & 23.18 & \textbf{23.40} & 22.41 & \textbf{22.59} & 21.81 & \textbf{21.99} & 21.33 & \textbf{21.47} \bigstrut[t]\\
          & 0.853 & 0.883 & 0.761 & 0.786 & 0.681 & \textbf{0.706} & 0.612 & \textbf{0.650} & 0.559 & \textbf{0.595} & 0.510 & \textbf{0.546} & 0.468 & \textbf{0.500} \\
    \hline
    \end{tabular}%
  \label{tab:result_natural}%
\end{table*}%

\subsection{Impact of the Normalization Parameter}\label{ssec:result_gamma}

In this experiment, we perform denoising on synthetic images with graph Laplacian regularization to examine the effects of choosing different $\gamma$'s. 
For testing, we used two $100\times 100$ synthetic images, as shown in the first column of Table\,\ref{tab:exp_gamma}. 
The image {\it Glow} was generated by a 2D circularly symmetric Gaussian with a standard deviation equal to 15 and mean located at the image center, and the binary image {\it Disk} has a white circular region of radius 15 on a black background.

We applied our method on the \emph{noiseless} versions of the \emph{Glow} and \emph{Disk} images. 
To be precise, we call this process \emph{filtering} rather than denoising in this experiment. 
For the {\it Glow} image, we treated it as a $100\times 100$ pixel patch, and let its similar patch be itself only. At each iteration, the result from the previous iteration, {\it i.e.}, ${\bf z}_0$ in \eqref{eq:prob_constrained}, was used to construct the graph Laplacian ${\bf L}$ for the current iteration. 
We set $\sigma_g^2=0$ and $\sqrt{\beta_{\cal G}}=0.02$ when computing $\{{\bf f}_1^\star, {\bf f}_2^\star, {\bf f}_3^\star\}$ in \eqref{eq:fea_spat} and \eqref{eq:fea_inten_2}. 
For a unified filtering strength, we solved \eqref{eq:prob_constrained} with fixed $\sigma_{\cal I}^2=1$ and $C_{\cal I}=1$. 
To see the impact of different $\gamma$'s, we set $\gamma$ to $\{0,1,2\}$ and filtered the image for 40 iterations. Similar filtering was also applied to the \emph{Disk} image.

Table\;\ref{tab:exp_gamma} shows the filtered images where they are slightly enhanced for better visualization. 
We observe that:
\begin{enumerate}[(i)]
\item{When $\gamma=0$, edges are well preserved---see the filtered {\it Disk} with $\gamma=0$. However, false edges are also created due to backward diffusion; {\it e.g.}, {\it Glow} is sharpened and has concentric circles after filtering with $\gamma=0$.}
\item{When $\gamma=1$, the filtered results are similar to the effects of TV denoising, which neither sharpen the image nor eliminate the transitions.}
\item{When $\gamma=2$, forward diffusion dominates and edges are smeared---see the filtered {\it Disk} with $\gamma=2$.}
\end{enumerate}
If the type of image to be denoised is known \textit{a priori}, then users can adjust the normalization parameter $\gamma$ accordingly, so as to achieve satisfactory denoising performance.
%
\begin{figure*}[!t]
\centering
    \stackunder[5pt]{\includegraphics[width=80pt]{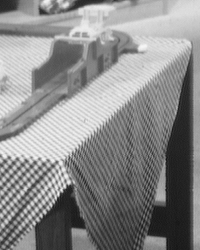}\hspace{2pt}\includegraphics[width=80pt]{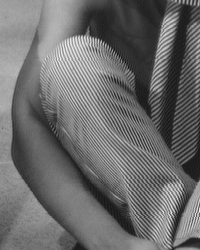}}{Original}\hspace{8pt}
    \stackunder[5pt]{\includegraphics[width=80pt]{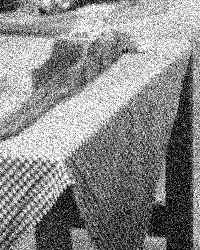}\hspace{2pt}\includegraphics[width=80pt]{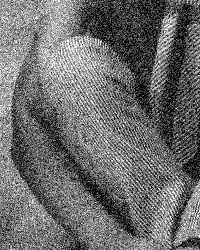}}{Noisy, 16.09\,dB}\hspace{8pt}
    \stackunder[5pt]{\includegraphics[width=80pt]{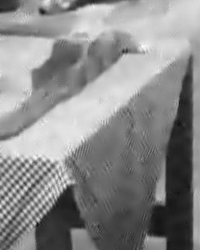}\hspace{2pt}\includegraphics[width=80pt]{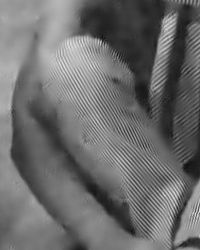}}{K-SVD, 26.84\,dB}\vspace{6pt}
    \stackunder[5pt]{\includegraphics[width=80pt]{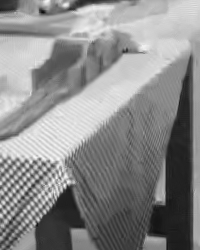}\hspace{2pt}\includegraphics[width=80pt]{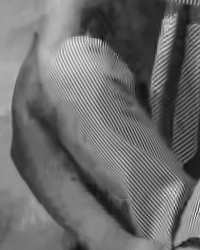}}{BM3D, 27.99\,dB}\hspace{8pt}
    \stackunder[5pt]{\includegraphics[width=80pt]{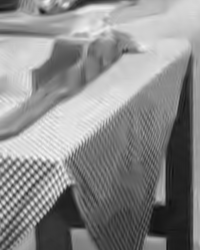}\hspace{2pt}\includegraphics[width=80pt]{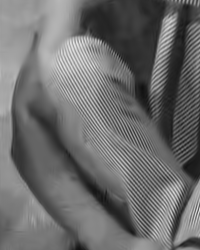}}{PLOW, 28.11\,dB}\hspace{8pt}
    \stackunder[5pt]{\includegraphics[width=80pt]{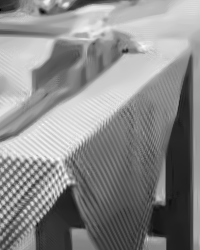}\hspace{2pt}\includegraphics[width=80pt]{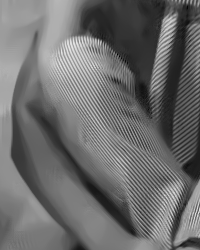}}{OGLR, 28.35\,dB}
\caption{Denoising of the natural image {\it Barbara}, where the original image is corrupted by AWGN with $\sigma_{\cal I}=40$. Two cropped fragments of each image are presented for comparison.}
\label{fig:result_barbara}
\end{figure*}

\subsection{Denoising of Natural Images}
\label{ssec:denoi_natural}

We next evaluate our OGLR algorithm using natural images:
four 512$\times$512 benchmark images (in grayscale)---{\it Lena}, {\it Barbara}, {\it Peppers}, and {\it Mandrill}.\footnote{Available at \url{http://sipi.usc.edu/database/}} 
The test images were corrupted by i.i.d. AWGN with standard deviation $\sigma_{\cal I}$ ranging from 10 to 70. 
We compared OGLR with three recent methods: K-SVD denoisng \cite{elad06}, BM3D \cite{dabov07}, and PLOW \cite{chatterjee12}.

For the natural images, the normalization parameter $\gamma$ was empirically set to be 0.6 for a reasonable trade-off between forward and backward diffusion. 
Depending on different noise variances $\sigma_{\cal I}^2$, we adjusted the patch length $\sqrt{M}$ from 7 to 22, adjusted the cluster size $K$ from 5 to 50, and adjusted $N_S$, the spacing between neighboring target patches, from 2 to 6. 
When computing graph weights in \eqref{eq:kernel}, $\epsilon$ was empirically set to be $4\%$ of the sum of $\sigma_{\cal I}^2$ and the maximum intensity difference of ${\bf z}_0$, and the threshold $r$ was chosen such that each vertex of graph $\mathcal{G}$ had at least 4 edges. 
We also set $\beta$ in \eqref{eq:G_atom} be $10^{-12}$---a very small value.
We ran OGLR and the competing methods over 5 independent noise realizations. 
For each $\sigma_{\cal I}$, the averaged objective performance, in terms of PSNR (in dB) and the SSIM index \cite{wang04}, are tabulated in Table\;\ref{tab:result_natural}.

From Table\;\ref{tab:result_natural}, we see that OGLR shows a performance close to that of BM3D. 
For the images {\it Barbara} and {\it Mandrill} with $\sigma_{\cal I}=40$, OGLR outperforms BM3D by up to 0.3\,dB. We also see that OGLR performs better than K-SVD and PLOW in most cases. 
Fig.\,\ref{fig:result_lena} shows two fragments of the image {\it Lena}, where the original fragments and the noisy versions (with $\sigma_{\cal I}=40$), accompanied by the denoised results, are presented for comparison. 
With similar settings, Fig.\,\ref{fig:result_barbara} shows different versions of two fragments of the image {\it Barbara}. 
We see that, compared to the other methods, our OGLR not only 
provides well-preserved textures, but also recovers flat regions faithfully, leading to a natural and satisfactory appearance.

\subsection{Denoising of Piecewise Smooth Images}\label{ssec:denoi_pws}
Both our analysis (Section\,\ref{ssec:pws}) and experimentation (Section\,\ref{ssec:result_gamma}) imply the effectiveness of graph Laplacian regularization for piecewise smooth images when parameter $\gamma$ is small. 
We herein evaluate OGLR on denoising of depth images---a class of grayscale images with piecewise smooth characteristics. 
Five benchmark depth images---{\it Cones}, {\it Teddy}, {\it Art}, {\it Moebius} and {\it Aloe}---were used.\footnote{Available at \url{http://vision.middlebury.edu/stereo/data/}} 

We compared OGLR with BM3D \cite{dabov07} and NLGBT \cite{hu13}. 
Note that NLGBT is a graph-based approach dedicated to depth image denoising with state-of-the-art performance. 
In this experiment, we set $\gamma=0$. 
The test images were corrupted by i.i.d. AWGN, with $\sigma_{\cal I}$ ranging from 10 to 50, then recovered with OGLR and the competing methods. 
The average objective performance of 5 independent noise realizations are presented in Table\;\ref{tab:result_pws}. First, we see that, NLGBT performed much better than BM3D. Moreover, among the three methods, our OGLR produced the best objective results in most cases, and outperformed NLGBT by up to 1.6\,dB ({\it Art}, $\sigma_{\cal I}=10$).

\begin{figure*}[!t]
\centering
    {\includegraphics[width=85pt]{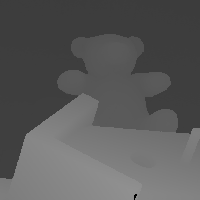}}\hspace{8pt}
    {\includegraphics[width=85pt]{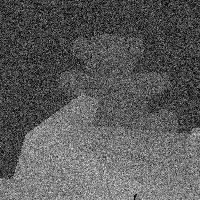}}\hspace{8pt}
    {\includegraphics[width=85pt]{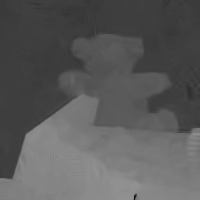}}\hspace{8pt}
    {\includegraphics[width=85pt]{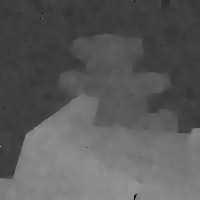}}\hspace{8pt}
    {\includegraphics[width=85pt]{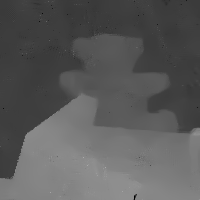}}\\\vspace{2pt}
    \stackunder[5pt]{\includegraphics[width=85pt]{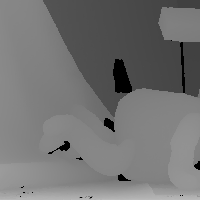}}{Original}\hspace{8pt}
    \stackunder[5pt]{\includegraphics[width=85pt]{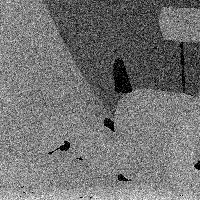}}{Noisy, 18.60\,dB}\hspace{8pt}
    \stackunder[5pt]{\includegraphics[width=85pt]{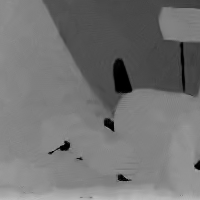}}{BM3D, 33.20\,dB}\hspace{8pt}
    \stackunder[5pt]{\includegraphics[width=85pt]{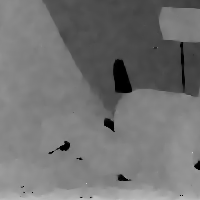}}{NLGBT, 33.94\,dB}\hspace{8pt}
    \stackunder[5pt]{\includegraphics[width=85pt]{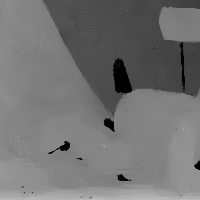}}{OGLR, 34.55\,dB}
\caption{Denoising of the depth image {\it Teddy}, where the original image is corrupted by AWGN with $\sigma_{\cal I}=30$. Two cropped fragments of each image are presented for comparison.}
\label{fig:result_teddy}
\end{figure*}
%

%
\begin{table*}[htbp]
  \centering\footnotesize
  \caption{Depth Image Denoising with OGLR: Performance Comparisons in
Average PSNR (Top, in dB) and SSIM Index (Bottom). In Each Cell, Results of Three Denoising Methods are Presented. Left: BM3D \cite{dabov07}. Middle: NLGBT \cite{hu13}. Right: OGLR (Proposed). The Best Results Among the Three Methods are Highlighted in Each Cell.}
    \begin{tabular}{|c||c:c:c|c:c:c|c:c:c|c:c:c|c:c:c|}
    \hline
    \multirow{2}[4]{*}{\textbf{Image}} & \multicolumn{15}{l|}{\hspace{-2pt}$\left[\begin{array}{c;{2pt/2pt}c;{2pt/2pt}c}{\textrm{BM3D}}&{\textrm{NLGBT}}&{\textrm{OGLR}}\end{array}\right]_{{\phantom{I}}_{\phantom{I}}}$\hspace{64pt}\textbf{\bf $\phantom{\hat{\hat{I}}}$Standard Deviation $\sigma_{\cal I}\phantom{\hat{\hat{I}}}$}} \\
\cline{2-16}          & \multicolumn{3}{c|}{\vspace{-2.7pt}$\phantom{\hat{I}}\mathop{\textrm{10}}\limits_{\phantom{.}}\phantom{\hat{I}}$} & \multicolumn{3}{c|}{20} & \multicolumn{3}{c|}{30} & \multicolumn{3}{c|}{40} & \multicolumn{3}{c|}{50} \\
    \hhline{================}
    \multirow{2}[2]{*}{\textit{Cones}} & 40.40 & 42.19 & \textbf{42.93} & 35.17 & 36.63 & \textbf{37.39} & 32.57 & 33.45 & \textbf{34.08} & 31.01 & 31.36 & \textbf{31.78} & 29.62 & 30.01 & \textbf{30.36} \bigstrut[t]\\
          & 0.983 & \textbf{0.987} & \textbf{0.987} & 0.960 & 0.966 & \textbf{0.968} & 0.935 & 0.942 & \textbf{0.944} & 0.912 & \textbf{0.926} & 0.922 & 0.898 & \textbf{0.913} & 0.900 \\
    \hline
    \multirow{2}[2]{*}{\textit{Teddy}} & 41.17 & 41.80 & \textbf{42.80} & 35.94 & 36.84 & \textbf{37.73} & 33.16 & 33.85 & \textbf{34.52} & 31.32 & 31.65 & \textbf{32.20} & 29.73 & 30.26 & \textbf{30.70} \bigstrut[t]\\
          & 0.985 & 0.985 & \textbf{0.986} & 0.967 & \textbf{0.968} & \textbf{0.968} & 0.948 & \textbf{0.949} & 0.947 & 0.927 & \textbf{0.937} & 0.929 & 0.919 & \textbf{0.928} & 0.910 \\
    \hline
    \multirow{2}[2]{*}{\textit{Art}} & 40.04 & 41.34 & \textbf{42.98} & 35.47 & 36.13 & \textbf{37.33} & 33.21 & 33.36 & \textbf{34.27} & 31.60 & 31.61 & \textbf{32.15} & 30.36 & 30.45 & \textbf{30.82} \bigstrut[t]\\
          & 0.983 & 0.986 & \textbf{0.988} & 0.959 & 0.963 & \textbf{0.967} & 0.934 & 0.937 & \textbf{0.944} & 0.907 & 0.920 & \textbf{0.922} & 0.891 & \textbf{0.906} & 0.898 \\
    \hline
    \multirow{2}[2]{*}{\textit{Moebius}} & 42.03 & 42.58 & \textbf{43.31} & 37.15 & 37.63 & \textbf{38.36} & 34.70 & 34.89 & \textbf{35.35} & 33.09 & 33.13 & \textbf{33.19} & 31.75 & \textbf{31.98} & 31.94 \bigstrut[t]\\
          & 0.983 & 0.984 & \textbf{0.985} & \textbf{0.962} & \textbf{0.962} & \textbf{0.962} & \textbf{0.940} & \textbf{0.940} & 0.938 & 0.918 & \textbf{0.929} & 0.917 & 0.911 & \textbf{0.922} & 0.898 \\
    \hline
    \multirow{2}[2]{*}{\textit{Aloe}} & 40.30 & 41.37 & \textbf{42.86} & 35.66 & 36.25 & \textbf{37.47} & 33.31 & 33.45 & \textbf{34.53} & 31.73 & 31.68 & \textbf{32.56} & 30.58 & 30.62 & \textbf{31.18} \bigstrut[t]\\
          & 0.984 & 0.986 & \textbf{0.988} & 0.962 & 0.965 & \textbf{0.968} & 0.938 & 0.941 & \textbf{0.946} & 0.913 & 0.925 & \textbf{0.928} & 0.899 & \textbf{0.913} & 0.907 \\
    \hline
    \end{tabular}%
  \label{tab:result_pws}%
\end{table*}%

\begin{figure*}[t]
\centering
    {\includegraphics[width=85pt]{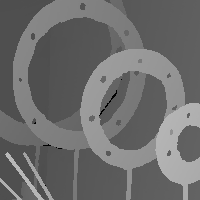}}\hspace{8pt}
    {\includegraphics[width=85pt]{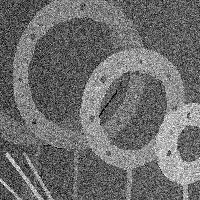}}\hspace{8pt}
    {\includegraphics[width=85pt]{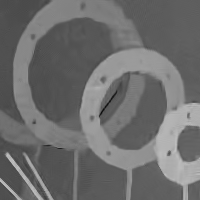}}\hspace{8pt}
    {\includegraphics[width=85pt]{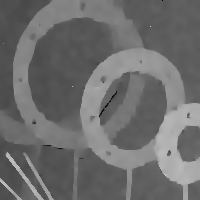}}\hspace{8pt}
    {\includegraphics[width=85pt]{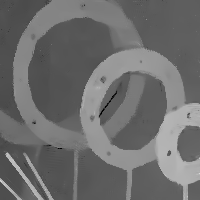}}\\\vspace{2pt}
    \stackunder[5pt]{\includegraphics[width=85pt]{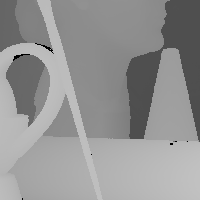}}{Original}\hspace{8pt}
    \stackunder[5pt]{\includegraphics[width=85pt]{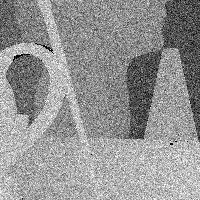}}{Noisy, 18.60\,dB}\hspace{8pt}
    \stackunder[5pt]{\includegraphics[width=85pt]{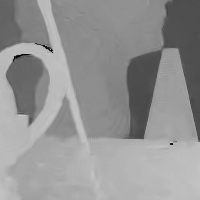}}{BM3D, 33.26\,dB}\hspace{8pt}
    \stackunder[5pt]{\includegraphics[width=85pt]{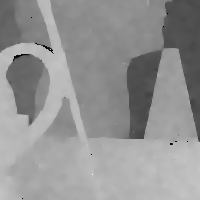}}{NLGBT, 33.41\,dB}\hspace{8pt}
    \stackunder[5pt]{\includegraphics[width=85pt]{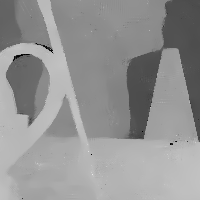}}{OGLR, 34.32\,dB}
\caption{Denoising of the depth image {\it Art}, where the original image is corrupted by AWGN with $\sigma_{\cal I}=30$. Two cropped fragments of each image are presented for comparison.}
\label{fig:result_art}
\end{figure*}

Visual comparisons are also shown in Fig.\;\ref{fig:result_teddy} and Fig.\;\ref{fig:result_art}, where different versions of fragments---original, noise-corrupted with $\sigma_{\cal I}=30$, and denoised---of the images \emph{Teddy} and \emph{Art} are presented, respectively. 
Compared to BM3D, NLGBT provided sharper transitions, though it failed to remove all the noise. 
In contrast, OGLR produced sharp edges while preserving the smoothness within each region. 

We note that on a desktop computer with an Intel Core i7 CPU, our brute-force MATLAB implementation of OGLR takes about 2 minutes to denoise a 256$\times$256 image with $\sigma_{\cal I}=30$. Its running time can be further reduced by having a more advanced implementation.

%

%% file: conclude.tex
The graph Laplacian regularizer is a popular recent prior to regularize inverse imaging problems. 
In this paper, to study in-depth the mechanisms and implications of graph Laplacian regularization, we regard a neighborhood graph as a discretization of a Riemannian manifold, and show convergence of the graph Laplacian regularizer to its continuous-domain counterpart. 
We then derive the optimal graph Laplacian regularizer for image denoising, assuming non-local self-similarity. 
To explain the behavior of graph Laplacian regularization, we interpret it as an anisotropic diffusion scheme in the continuous domain, and delineate its relationship to the well-known total variation (TV) prior.
Our developed denoising algorithm, optimal graph Laplacian regularization (OGLR) for denoising, produces competitive results for natural images compared to state-of-the-art methods, and out-performs them for piecewise smooth images.

%% file: conv_proof.tex
\begin{proof}
Let $\mathcal{M}$ be a 2D Riemannian manifold embedded in $N$ dimensional ambient space through the continuous embedding ${\Phi}:\mathcal{M}\mapsto\mathbb{R}^N$. 
Specifically,
\begin{equation}\label{eq:emb}
{\Phi}\hspace{-2pt}:\hspace{-2pt}({\sigma _1},{\sigma _2})\hspace{-2pt}\mapsto\hspace{-2pt}\left( {{f_1}({\sigma _1},{\sigma _2}),{f_2}({\sigma _1},{\sigma _2}), \ldots ,{f_N}({\sigma _1},{\sigma _2})} \right),
\end{equation}
where $(\sigma_1,\sigma_2)$ are the global coordinates of $\mathcal{M}$. 
Under embedding $\Phi$, the induced metric of $\mathcal{M}$ in $\mathbb{R}^{N}$ can be pulled back (as done in \cite{sochen98}), which is the matrix ${\bf{G}}$ \eqref{eq:metric}.

Then we relate the sampling positions in $\Gamma$ to a probability density function (PDF) defined on manifold $\mathcal{M}$. 
Let the one-to-one mapping ${\Psi}:\mathcal{M}\mapsto \Omega$ be
\begin{equation}
{\Psi}:({\sigma _1},{\sigma _2}) \mapsto (x = {\sigma _1},y = {\sigma _2}).
\end{equation}
Then let the function $p(x,y): \Omega\rightarrow \mathbb{R}$ be
\begin{equation}\label{eq:prob}
p(x,y) = 1/(\left| \Omega  \right|\sqrt {\det {\bf{G}}}),
\end{equation}
where $\left| \Omega  \right|$ denotes the area of $\Omega$. 
Through mapping ${\Psi}$, a function $p^\mathcal{M}({\sigma _1},{\sigma _2}):\mathcal{M}\mapsto[0,+\infty)$, is obtained, where $p^\mathcal{M}({\sigma _1},{\sigma _2})=p({\Psi }({\sigma _1},{\sigma _2}))$. 
Because of ${\Psi}$, $p^\mathcal{M}$ and $p$ have same functional form, though they are defined in different domains. 
Moreover, from \eqref{eq:prob}, $p^\mathcal{M}$ is a PDF on $\mathcal{M}$ because
\begin{equation}
\int_\mathcal{M} {{p^\mathcal{M}}({\sigma _1},{\sigma _2})} \hspace{1.5pt} dV = \int_{\Omega} {p\sqrt {\det {\bf{G}}} } \hspace{1.5pt}\,d{\bf{s}} = 1,
\end{equation}
where $dV\hspace{-4pt}=\hspace{-4pt} {\sqrt{\det {\bf{G}}} } \hspace{1.5pt}\,d{\bf{s}}$ is the natural volume element of $\mathcal{M}$.

For any sub-domain $\mathcal{M}'\subseteq \mathcal{M}$, its counterpart on $\Omega$ is $\Omega'=\{{\Psi}(\sigma_1,\sigma_2)|(\sigma_1,\sigma_2)\in \mathcal{M}'\}\subseteq \Omega$. 
Assume the tuple $(\widehat{\sigma}_1,\widehat{\sigma}_2)$ is a 2D random variable on $\mathcal{M}$ with density function $p^\mathcal{M}$. 
Then $(\widehat{x},\widehat{y})={\Psi}(\widehat{\sigma}_1,\widehat{\sigma}_2)$ is the corresponding 2D random variable on $\Omega$. 
Since the probability
\begin{equation}
\begin{split}
&Pr((\widehat{x},\widehat{y})\in\Omega')=Pr((\widehat{\sigma}_1,\widehat{\sigma}_2)\in \mathcal{M}')\\
&=\int_{\mathcal{M}'} {{p^\mathcal{M}}({\sigma _1},{\sigma _2})}\, dV=\int_{\Omega'} {p\sqrt {\det {\bf{G}}} } \hspace{1.5pt} d{\bf{s}}=\frac{|\Omega'|}{|\Omega|},
\end{split}
\end{equation}
$(\widehat{x},\widehat{y})$ follows \emph{uniform distribution} on $\Omega$. 
As a result, the set $\Gamma$ containing uniformly distributed positions in $\Omega$, is generated as follows: $M$ positions on manifold $\mathcal{M}$ are drawn independently according to $p^\mathcal{M}$ and then are mapped to $\Omega$ through $\Psi$.

With the above settings, from \eqref{eq:sample}, \eqref{eq:weight}, and \eqref{eq:emb}, graph $\mathcal{G}$ is built upon $M$ samples from manifold $\mathcal{M}$, where these $M$ samples are uniformly disributed on $\Omega$ after being mapped by $\Psi$.
According to \cite{hein06} and \cite{hein07}, the discrete graph $\mathcal{G}$ is an approximation of the manifold $\mathcal{M}$. 
For a smooth function $u$ on $\Omega$, its counterpart on $\mathcal{M}$ is ${u^\mathcal{M}}({\sigma _1},{\sigma _2}) = u({\Psi }(\sigma_1,\sigma_2))$ and its discretized version is ${\bf{u}}$. 
According to \cite{hein06}, if $\mathcal{M}$ is a smooth compact manifold with a boundary, $u^{\mathcal{M}}$ belongs to the class of $\kappa$-H\"{o}lder functions with $\kappa\ge3$ and the weight parameter $\epsilon=O\left(M^{-\frac{\kappa}{4(\kappa+2)}}\right)$. Then we have:\footnote{We refer readers to \cite{hein06} for a uniform convergence result on a more general basis and its corresponding assumptions on $\mathcal{M}$, $u^\mathcal{M}$ and $\epsilon$.}
\begin{equation}\label{eq:unif_conv}
\sup\hspace{-1pt}\left|\hspace{-1pt}{\frac{{c{M^{2\gamma  - 1}}}}{{{\epsilon^{4(1 - \gamma )}}(M\hspace{-2pt}-\hspace{-5pt}1)}}{S_{\cal G}}({{\bf{u}}})\hspace{-2pt}-\hspace{-2pt}{S_\Delta }({u^{\cal M}})}\hspace{-1pt}\right|\hspace{-2pt}\hspace{-1pt}=\hspace{-1pt}O\hspace{-2pt}\left(\hspace{-2pt}{{M^{ - \frac{\kappa }{{4(\kappa  + 2)}}}}}\hspace{-2pt}\right),
\end{equation}
where $c$ is a constant that only depends on $C_r$. 
The functional $S_{\Delta}$ is induced by the $2(1-\gamma)$-th weighted \emph{Laplace-Beltrami operator} for $\kappa$-H\"{o}lder functions on manifold $\mathcal{M}$. 
It is
\begin{equation}\label{eq:beltrami}
\begin{split}
&{S_{{\Delta}}}({u^\mathcal{M}}) = \int_\mathcal{M} {\left\langle {\nabla {u^\mathcal{M}},\nabla {u^\mathcal{M}}} \right\rangle } {({p^\mathcal{M}})^{2(1-\gamma)}}\,dV\\
&\xlongequal{\eqref{eq:prob}} \int_\Omega  {{{({{\bf{G}}^{ - 1}}\nabla u)}^{\rm{T}}}{\bf{G}}({{\bf{G}}^{ - 1}}\nabla u){p^{2(1 - \gamma )}}\sqrt {\det {\bf{G}}} }\hspace{1.5pt} d{\bf{s}}\\
&= {{\left| \Omega  \right|}^{-2(1-\gamma)}}\hspace{-5pt}\int_\Omega  {{{\nabla u}^{\rm{T}}}{{\left( {{{\left( {\det {\bf{G}}} \right)}^{\frac{1}{2} - \gamma }}{\bf{G}}} \right)}^{ - 1}}\nabla u}\hspace{2pt}d{\bf{s}},
\end{split}
\end{equation}
which equals ${\left| \Omega  \right|^{ -2(1-\gamma)}}{S_\Omega }(u)$. 
From \eqref{eq:unif_conv} and \eqref{eq:beltrami}, \eqref{eq:conv} is readily obtained by weakening the uniform convergence of \eqref{eq:unif_conv} to point-wise convergence.
\end{proof}

%% file: bio.tex
\begin{biography}[{\includegraphics[width=1in,height=1.25in,clip,keepaspectratio]{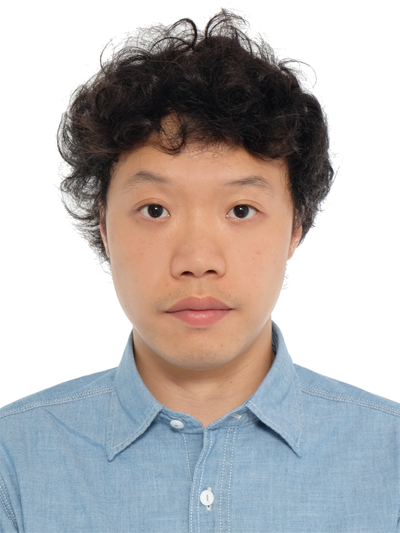}}]{Jiahao Pang} (S'13---M'16) 
received the B.Eng. degree from South China University of Technology, Guangzhou, China, in 2010, and the M.Sc. and Ph.D. degrees from the Hong Kong University of Science and Technology, Hong Kong, in 2011 and 2016, respectively. 
He also conducted his research in National Institute of Informatics in Tokyo, Japan (2014--2016). 
He is currently a researcher with SenseTime Group Limited, Hong Kong. His research interests include image/video processing and graph signal processing.
\end{biography}

\begin{biography}[{\includegraphics[width=1in,height=1.25in,clip,keepaspectratio]{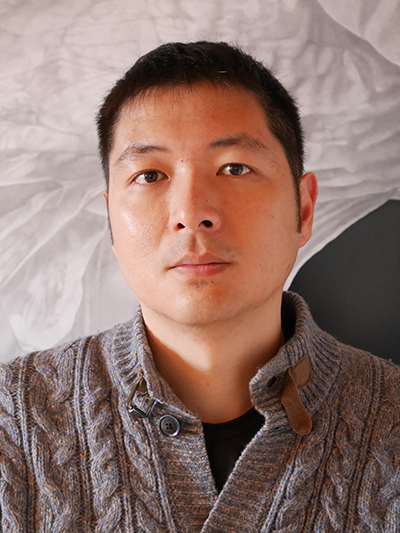}}]{Gene Cheung} (M'00---SM'07)
received the B.S. degree in electrical engineering from Cornell University in 1995, and the M.S. and Ph.D. degrees in electrical engineering and computer science from the University of California, Berkeley, in 1998 and 2000, respectively. 

He was a senior researcher in Hewlett-Packard Laboratories Japan, Tokyo, from 2000 till 2009. 
He is now an associate professor in National Institute of Informatics in Tokyo, Japan. 
He has been an adjunct associate professor in the Hong Kong University of Science \& Technology (HKUST) since 2015.

His research interests include 3D image processing, graph signal processing, and signal processing for sleep analysis. 
He has served as associate editor for IEEE Transactions on Multimedia (2007--2011), DSP Applications Column in IEEE Signal Processing Magazine (2010--2014) and SPIE Journal of Electronic Imaging (2014--2016). 
He currently serves as associate editor for IEEE Transactions on Image Processing (2015--present), IEEE Transactions on Circuits and Systems for Video Technology (2016--present) and APSIPA Journal on Signal \& Information Processing (2011--present), and as area editor for EURASIP Signal Processing: Image Communication (2011--present). 
He is a distinguished lecturer in APSIPA (2016--2017). 
He served as a member of the Multimedia Signal Processing Technical Committee (MMSP-TC) in IEEE Signal Processing Society (2012--2014), and a member of the Image, Video, and Multidimensional Signal Processing Technical Committee (IVMSP-TC) (2015--2017). 
He has also served as technical program co-chair of International Packet Video Workshop (PV) 2010 and IEEE International Workshop on Multimedia Signal Processing (MMSP) 2015, and symposium co-chair for CSSMA Symposium in IEEE GLOBECOM 2012. 
He is a co-author of the best student paper award in IEEE Workshop on Streaming and Media Communications 2011 (in conjunction with ICME 2011), ICIP 2013 and IVMSP 2016, best paper runner-up award in ICME 2012, and best paper finalists in ICME 2011, ICIP 2011 and ICME 2015.
\end{biography}